\documentclass{article}

\usepackage{graphicx}
\usepackage{caption}
\usepackage{booktabs}
\usepackage{multirow}
\usepackage{siunitx}
\usepackage{amsmath, amssymb}
\usepackage{subcaption}
\usepackage{booktabs}
\usepackage{makecell}
\usepackage[table]{xcolor}
\usepackage[preprint]{corl_2026} 

\makeatletter
\renewcommand{\@conferencelocation}{Austin, Texas, USA}
\makeatother

\usepackage{wrapfig}

\usepackage{enumitem}
\setlist[itemize]{topsep=2pt, partopsep=0pt, parsep=2pt, itemsep=2pt, leftmargin=1.6em}
\setlist[enumerate]{topsep=2pt, partopsep=0pt, parsep=2pt, itemsep=2pt, leftmargin=1.8em}
\makeatletter
\renewcommand\paragraph{%
  \@startsection{paragraph}{4}{\z@}%
  {1.0ex \@plus 0.4ex \@minus 0.2ex}%
  {-0.6em}%
  {\normalfont\normalsize\bfseries}}
\renewcommand\section{%
  \@startsection{section}{1}{\z@}%
  {2.2ex \@plus 0.6ex \@minus 0.2ex}%
  {1.0ex \@plus 0.2ex}%
  {\normalfont\Large\bfseries}}
\renewcommand\subsection{%
  \@startsection{subsection}{2}{\z@}%
  {1.6ex \@plus 0.4ex \@minus 0.2ex}%
  {0.6ex \@plus 0.1ex}%
  {\normalfont\large\bfseries}}
\makeatother
\begin{document}

\title{WoVR: World Models as Reliable Simulators for Post-Training VLA Policies with RL}

\author{
\textbf{Zhennan Jiang}{\textsuperscript{\textnormal{1,3,4,*}}} \quad
\textbf{Shangqing Zhou}{\textsuperscript{\textnormal{1,3,*}}} \quad
\textbf{Yutong Jiang}{\textsuperscript{\textnormal{1}}} \quad
\textbf{Zefang Huang}{\textsuperscript{\textnormal{4}}} \\
\textbf{Mingjie Wei}{\textsuperscript{\textnormal{4}}} \quad
\textbf{Yuhui Chen}{\textsuperscript{\textnormal{1,3}}} \quad
\textbf{Tianxing Zhou}{\textsuperscript{\textnormal{4}}} \quad
\textbf{Zhen Guo}{\textsuperscript{\textnormal{5}}} \quad 
\textbf{Hao Lin}{\textsuperscript{\textnormal{5}}} \\
\textbf{Quanlu Zhang}{\textsuperscript{\textnormal{5}}} \quad
\textbf{Yu Wang}{\textsuperscript{\textnormal{2}}} \quad
\textbf{Dongbin Zhao}{\textsuperscript{\textnormal{1,3,4, \dag}}} \quad
\textbf{Haoran Li}{\textsuperscript{\textnormal{1,3, \dag}}} \quad
\textbf{Chao Yu}{\textsuperscript{\textnormal{2, \dag}}} \\
\\[-2pt]
\textsuperscript{\textnormal{$*$}}Equal contribution \qquad
\textsuperscript{\textnormal{$\dagger$}}Corresponding authors \\[-2pt]
\textsuperscript{1}University of Chinese Academy of Sciences \quad
\textsuperscript{2}Tsinghua university \\
\textsuperscript{3}Institute of Automation, Chinese Academy of Sciences \\
\textsuperscript{4}Zhongguancun Academy  \quad
\textsuperscript{5}Infinigence AI 
\\[1.1em]
{\normalsize
\textcolor[HTML]{334155}{\textbf{Project page:}}\quad
\href{https://wovr-rlinf.github.io/}{\textcolor[HTML]{D12B83}{\texttt{https://wovr-rlinf.github.io/}}}
}
}

\makeatletter
\hypersetup{
  pdftitle={\@title},
  pdfauthor={Zhennan Jiang, Shangqing Zhou, Yutong Jiang, Zefang Huang, Mingjie Wei, Yuhui Chen, Tianxing Zhou, Zhen Guo, Hao Lin, Quanlu Zhang, Yu Wang, Dongbin Zhao, Haoran Li, Chao Yu}
}
\makeatother

\maketitle

\begin{abstract}
Reinforcement learning (RL) promises to unlock capabilities beyond imitation learning for Vision--Language--Action (VLA) models, but its requirement for massive real-world interaction prevents direct deployment on physical robots. 
Recent work attempts to use learned world models as simulators for policy optimization, yet closed-loop imagined rollouts inevitably suffer from hallucination and long-horizon error accumulation. 
Such errors not only degrade visual fidelity, but also mislead policy optimization by providing unreliable learning signals. 
We propose \textbf{WoVR}, a reliable world-model-based RL framework for post-training VLA policies. 
Instead of assuming a faithful world model, WoVR explicitly regulates how RL interacts with imperfect imagined dynamics. 
It improves rollout stability through a controllable action-conditioned video world model, reshapes imagined interaction to reduce effective error depth via Keyframe-Initialized Rollouts, and maintains policy--simulator alignment through World Model-Policy co-evolution.
Extensive experiments demonstrate that WoVR enables stable long-horizon imagined rollouts and effective policy optimization, achieving superior LIBERO performance and consistent real-world gains across multiple robotic platforms.
These results show that world models can serve as practical simulators for RL when hallucination is explicitly controlled.
\end{abstract}

\keywords{World Model, Reinforcement Learning, Vision-Language-Action} 

\section{Introduction}

Vision--Language--Action (VLA) models~\citep{kim2025fine, black2024pi0visionlanguageactionflowmodel, intelligence2025pi05visionlanguageactionmodelopenworld, li2025surveyvisionlanguageactionmodelsembodied, pei2026evolutionvla} have been increasingly adopted for robotic manipulation, where actions are generated end-to-end by conditioning on language instructions and visual observations. 
Most existing VLA systems are trained via imitation learning. 
While effective in many downstream tasks, this paradigm fundamentally limits the performance ceiling of VLA policies, as it is tightly constrained by the quality and coverage of demonstration data.

Recent VLA-RL methods~\citep{zang2025rlinfvlaunifiedefficientframework, liu2026rlbringvlageneralization, chen2026pitextttrlonlinerlfinetuning, chen2025conrft, lu2025vlarlmasterfulgeneralrobotic, li2025grrlgoingdexterousprecise,liu2026lifelongrft} improve beyond imitation learning~\citep{lei2025rl100performantroboticmanipulation}. 
However, applying RL to real-world VLA policies presents a fundamental dilemma: off-policy methods are generally more sample-efficient but often suffer from distribution shift and training instability, whereas on-policy updates \citep{schulman2017proximal, shao2024deepseekmathpushinglimitsmathematical, guo2025deepseek} require massive parallel environment interaction for stable and efficient training, making them impractical for real robots due to costly data collection~\citep{cui2024gapartmanip,luo2025precise}.
Although simulation-based alternatives have been explored~\citep{chen2026pitextttrlonlinerlfinetuning, li2025simplevla, liu2025fetchbot}, aligning simulators with real-world dynamics remains highly challenging. These constraints motivate replacing real-environment interaction with a learned world model that serves as a simulator for policy optimization.

This direction has become increasingly viable with the emergence of large-scale generative video models~\citep{opensora, wan2025wanopenadvancedlargescale}.
Several works directly treat pretrained video generators as simulators and perform reinforcement learning entirely in imagination~\citep{zhu2025wmpoworldmodelbasedpolicy, li2025vlarftvisionlanguageactionreinforcementfinetuning}.
\textbf{However, learned world models are not faithful simulators.}
In this work, we define \emph{hallucination} as a systematic mismatch between imagined and real outcomes in closed-loop interaction. The world model may generate visually plausible yet physically incorrect rollouts, resulting in spurious success signals under the policy's actions.
In closed-loop autoregressive rollouts, prediction errors compound with horizon length due to:
\begin{itemize}
    \item Autoregressive feedback: the model conditions on its own generated frames, amplifying small early errors;
    \item Distribution shift: as the policy evolves, its action distribution drifts away from the data used to train the world model, increasing out-of-distribution prediction failures.
\end{itemize}

\begin{figure}[t]
  \centering
  \includegraphics[width=0.9\linewidth]{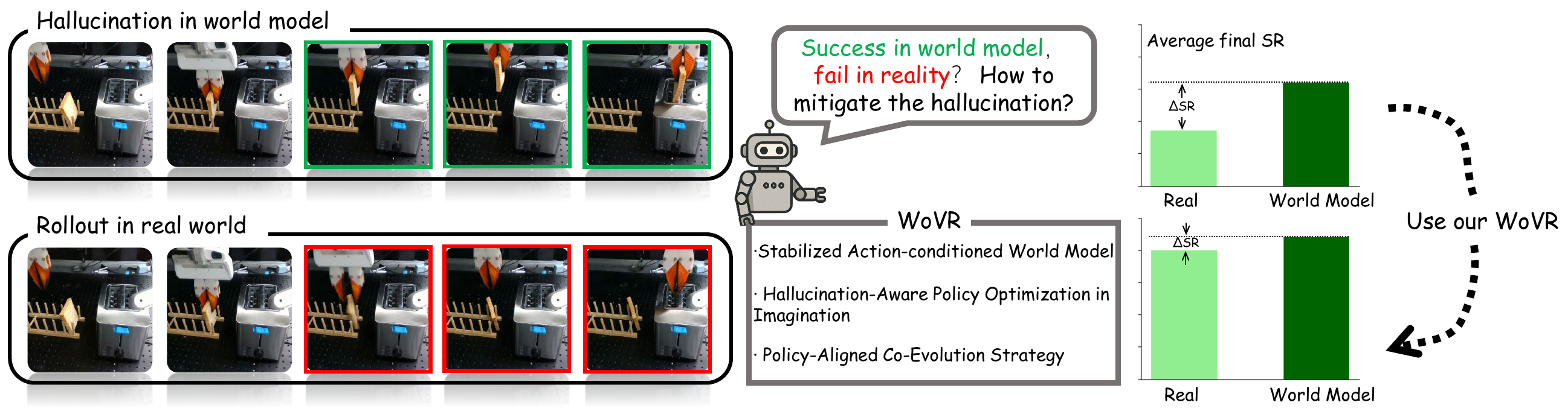} 
  \caption{\textbf{Hallucination in Closed-Loop World Model Rollouts.} The world model imagines a successful grasp (green frames), but real-world execution fails (red frames).To address this critical mismatch, we propose three hallucination-aware mechanisms.}
  \label{fig:hallucination}
\end{figure}

If hallucinated trajectories are directly used for policy optimization, RL is incentivized to exploit systematic model errors rather than true task progress.
This leads to a critical question:
\vspace{-0.6em}
\begin{center}
\emph{If world models inevitably hallucinate, how can RL remain reliable under imperfect imagined dynamics?}
\end{center}
\vspace{-0.6em}
We argue that using world models for RL is not primarily a modeling problem, but a reliability problem.
To make world-model-based RL reliable, one must control hallucination at three interconnected levels: controllable simulator design, reliable interaction protocol, and policy--model alignment. 
To this end, we propose \textbf{WoVR}, a \textbf{Wo}rld-model-based framework for post-training \textbf{V}ision--Language--Action policies with \textbf{R}einforcement Learning. 
Rather than assuming the learned world model to be a faithful simulator, WoVR explicitly regulates how reinforcement learning interacts with imperfect imagined dynamics. 
We first strengthen the simulator itself by constructing a rollout-stable, action-controllable video world model with stabilized autoregressive context modeling, reducing long-horizon drift and structural collapse. 
To mitigate long-horizon prediction-error accumulation, we further introduce \emph{Keyframe-Initialized Rollouts (KIR)}, which start imagined rollouts from task-critical states, shortening the effective prediction depth and reducing hallucination compounding.
Finally, as policy optimization shifts the action distribution and induces distribution mismatch between the policy and the world model, we introduce \emph{PACE}, a policy-aligned co-evolution strategy that restores alignment by iteratively refining the world model under the evolving policy distribution, without requiring continuous online supervision. 
Together, these components form a unified hallucination-aware reinforcement learning framework that enables reliable policy optimization in imagination. 
In summary, our contributions are as follows.
\begin{itemize}
    \item We identify hallucination under closed-loop imagined interaction as a fundamental reliability challenge in world-model-based RL for VLA, showing that autoregressive error accumulation and policy-induced distribution shift can systematically corrupt optimization signals.

    \item We propose WoVR, a hallucination-aware RL framework that jointly regulates controllable simulator design, reliable imagined interaction, and a policy-aligned co-evolution strategy, enabling stable on-policy optimization entirely in imagination.

    \item WoVR achieves state-of-the-art world-model quality with strong perceptual and temporal consistency while maintaining high rollout efficiency at 23 FPS. 
    More importantly, it consistently improves policy performance in both simulation and real-world deployment, boosting LIBERO success under different SFT initializations and improving success rates across multiple robotic platforms.
\end{itemize}

\begin{figure}[t]
  \centering
  \includegraphics[width=0.9\linewidth]{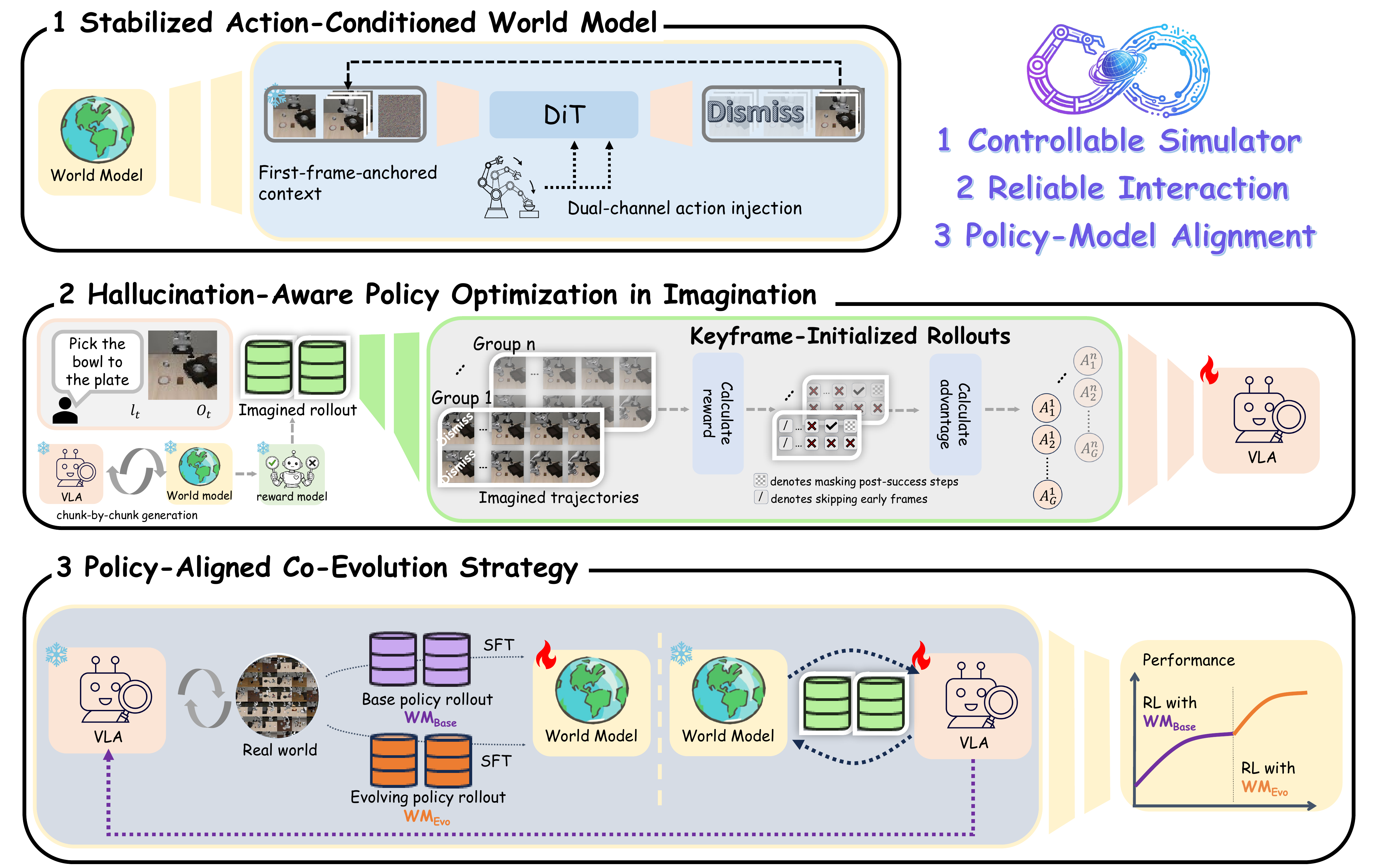}
  \caption{\textbf{Overview of WoVR}. WoVR builds a reliability-driven reinforcement learning framework entirely around the learned world model.}
  \label{fig:main}
\end{figure}

\vspace{-1.0em}
\section{Related Work}
\vspace{-0.6em}

\subsection{RL Fine-tuning for VLA Models}
\vspace{-0.45em}
On-policy reinforcement learning~\citep{zang2025rlinfvlaunifiedefficientframework,schulman2017proximal, zhang2026conformalizedppo} has been increasingly adopted to fine-tune VLA models beyond imitation learning~\citep{zang2025rlinfvlaunifiedefficientframework,liu2026rlbringvlageneralization}. However, directly transferring on-policy fine-tuning to real robots remains impractical, as such methods require large-scale parallel rollouts, repeated environment resets, and tightly coupled policy--environment interaction, which are difficult to support under real-world hardware. To mitigate this, some off-policy approaches~\citep{chen2025conrft, yuan2024policydecoratormodelagnosticonline} introduce offline data reuse or human intervention, but often suffer from limited scalability and performance degradation during online updates. An alternative direction builds large-scale real-robot infrastructures, yet existing systems~\citep{pan2026sopscalableonlineposttraining, zang2026rlinfuserunifiedextensiblerealworld} still cannot practically support fully on-policy algorithms at scale. These limitations suggest that the challenge of online RL for VLA is systemic rather than algorithmic, motivating world-model-based approaches that decouple policy optimization from real-world interaction.

\subsection{World Models for Policy Optimization}
\vspace{-0.45em}
A growing body of work has begun to incorporate learned world models into VLA policy optimization~\citep{quevedo2025worldgymworldmodelenvironment,jiang2025enerverseacenvisioningembodiedenvironments,jiang2025world4rldiffusionworldmodels}. One line of work uses world models primarily as reward or evaluation signals for post-training, where predicted future outcomes are used to construct preferences or guide policy improvement~\citep{fei2025srposelfreferentialpolicyoptimization,hung2025nora15visionlanguageactionmodeltrained,sun2026atomvlascalableposttrainingrobotic,chen2026tevir,liu2026deg}. 
Another line improves VLA policies by generating synthetic rollouts with learned world models, reducing reliance on costly real-world interaction~\citep{guo2025ctrlworldcontrollablegenerativeworld,yang2026riseselfimprovingrobotpolicy,guo2026vlawiterativecoimprovementvisionlanguageaction}. 
However, these works stop short of using world models as closed-loop RL simulators for policy optimization.

Prophet~\citep{zhang2025reinforcingactionpoliciesprophesying}, World-Env~\citep{xiao2025worldenvleveragingworldmodel} and WMPO~\citep{zhu2025wmpoworldmodelbasedpolicy} move in this direction by replacing real-environment interaction with imagined rollouts during reinforcement learning. Yet both largely treat the world model as a drop-in simulator, without explicitly addressing the central challenge of hallucinated dynamics: in closed-loop rollouts, prediction errors accumulate and can be exploited by the policy, ultimately misleading optimization.

\section{Methods}

We propose WoVR, a reliability-driven world-model-based reinforcement learning framework for post-training VLA policies. 
As illustrated in Fig.~\ref{fig:main}, WoVR treats the world model as a generative simulator and controls hallucination across three levels:
(1) Simulator-level control: we construct an action-controllable, rollout-stable video world model with dual-channel action injection and first-frame anchoring to suppress long-horizon drift. 
(2) Interaction-level reshaping: we redesign imagined interaction through Keyframe-Initialized Rollouts (KIR) to reduce effective error depth and prevent optimization on hallucinated success. 
(3) Alignment-level regulation: we introduce PACE, a policy--model co-evolution strategy that mitigates distribution shift by periodically aligning the world model with the evolving policy. 

\vspace{-1.0em}
\subsection{Stabilized Action-Conditioned World Model}
\vspace{-0.6em}
WoVR relies on a learned video world model as a generative simulator for closed-loop imagined interaction. 
However, long-horizon autoregressive generation can accumulate hallucinations, causing scene drift and appearance degradation.
We therefore design the world model to be both \emph{action-controllable} and \emph{rollout-stable}, so that the simulated dynamics remain consistent under iterative, policy-driven generation.

\begin{figure}[bhtp]
  \centering
  \includegraphics[width=0.9\linewidth]{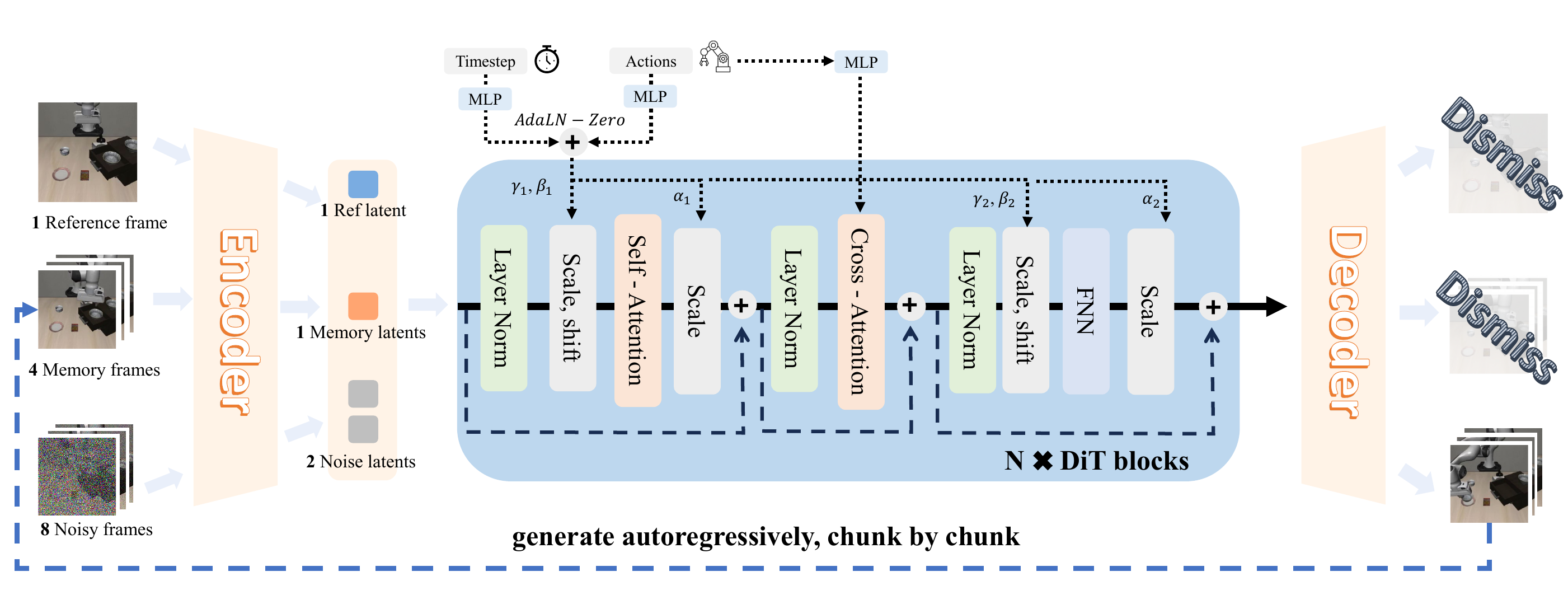} 
  \caption{\textbf{Architecture of the proposed action-conditioned world model.}}
  \label{fig:model_arch}
\end{figure}

Our world model is built on the Wan2.2-TI2V-5B video diffusion backbone~\citep{wan2025wanopenadvancedlargescale} and reformulated as an action-conditioned generator.
As shown in Fig.~\ref{fig:model_arch}, actions are injected through two complementary pathways: they modulate denoising features through timestep-conditioned normalization and replace text embeddings in cross-attention to provide global action context.
This \emph{dual-channel design} preserves the original DiT structure while enabling frame-level control.

To stabilize closed-loop rollout, we additionally use a \emph{first-frame--anchored} context.
At each autoregressive step, the model conditions on $[o_{0},\, o_{t-c:t}]$, combining the initial frame with recent memory frames from the previous chunk.
The fixed reference frame constrains global layout and appearance, while the memory frames preserve local dynamics, reducing drift and background collapse in long-horizon generation~\citep{shin2025motionstreamrealtimevideogeneration,yang2025longliverealtimeinteractivelong,tang2025hunyuangamecraft2instructionfollowinginteractivegame}.

During training, we apply \emph{noisy context augmentation}: non-reference context frames are mildly noised, while the first-frame anchor remains clean.
This makes the model robust to self-generated context at inference time and reduces brittle copying from previous predicted frames.

Together, dual-channel action conditioning, first-frame anchoring, and noisy context augmentation turn the video generator into a rollout-stable simulator for imagined RL.
Given anchored context and policy actions, the model autoregressively predicts and appends video chunks, producing long-horizon trajectories entirely in imagination.
More details are provided in Appendix~\ref{app:impl:world_model}.

When used as an RL simulator, the world model must also provide a reward signal. We support two modeling choices: a lightweight ResNet-based model for binary rewards, and a Qwen3-VL-based model for dense rewards. In practice, we find that the ResNet-based model is significantly more time-efficient while achieving comparable performance. Details are given in Appendix~\ref{sec:rm}.

\vspace{-1.0em}
\subsection{Hallucination-Aware Policy Optimization in Imagination}
\vspace{-0.6em}
WoVR optimizes the VLA policy by interacting with the learned world model, which serves as a generative simulator for closed-loop imagined rollouts. 
The key difficulty is that, in long-horizon rollouts starting from the initial state, world-model errors accumulate early and can eventually produce visually plausible but physically incorrect transitions and even spurious success signals.
If reinforcement learning naively trusts such rollouts, the policy is encouraged to optimize toward hallucinated outcomes rather than real task progress.

\begin{wrapfigure}{r}{0.6\linewidth}
  \vspace{-0.8em}
  \centering
  \includegraphics[width=\linewidth]{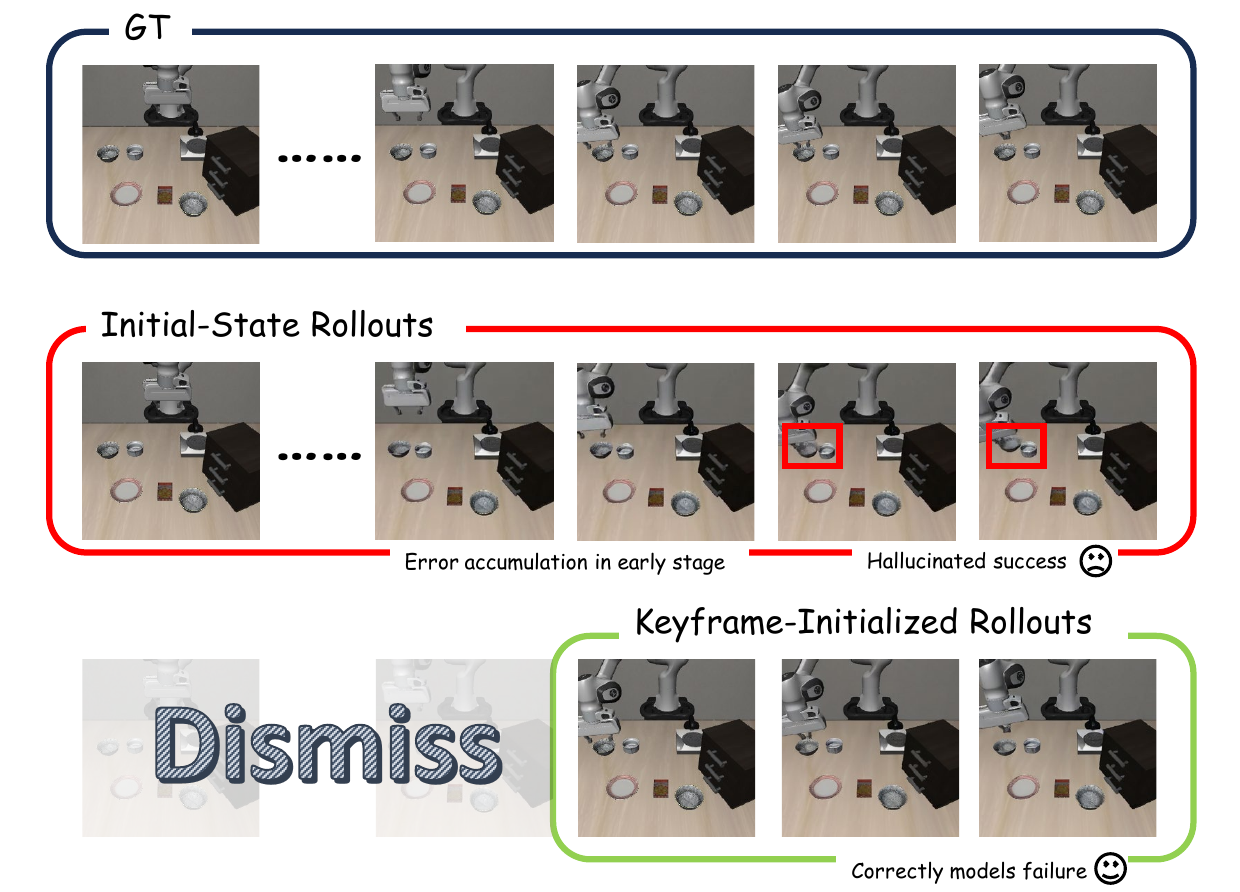}
  \caption{\textbf{Illustration of Keyframe-Initialized Rollouts (KIR).} Long-horizon rollouts accumulate prediction errors in early stages, leading to hallucinated success that contradicts the ground-truth failure. In contrast, KIR initialize rollouts near critical states, enabling physically consistent predictions that correctly model failure.}
  \label{fig:KIR}
  \vspace{-1.0em}
\end{wrapfigure}

To reduce the effective error depth of imagined interaction, we introduce Keyframe-Initialized Rollouts (KIR).
As illustrated in Figure~\ref{fig:KIR}, instead of always initializing rollouts from the episode start $o_0$, we initialize a portion of rollouts from keyframes $o_k$ that lie near task-critical intermediate states, especially failure states encountered by the current policy.
The motivation is that many decisive contacts and corrections happen locally around these states, whereas starting from $o_0$ forces the world model to predict a long prefix before reaching them, during which compounding errors can already derail the rollout.

We adopt GRPO to update the policy using imagined rollouts. The GRPO objective is defined as:
\begin{equation}
\begin{aligned}
J_{\mathrm{GRPO}}(\theta) = 
\mathbb{E}\Bigg[
\frac{1}{G}\sum_{i=1}^{G}\frac{1}{T_i^{\mathrm{valid}}}
\sum_{t=1}^{T_i^{\mathrm{valid}}}
\min\!\Big(
\rho_t^{(i)}(\theta)\hat{A}^{(i)}, 
\mathrm{clip}\!\big(\rho_t^{(i)}(\theta),\,1-\epsilon,\,1+\epsilon\big)\hat{A}^{(i)}
\Big)
\Bigg],
\end{aligned}
\end{equation}

where $G$ is the number of imagined trajectories, $\hat{A}^{(i)}$ is the group-relative advantage. and $T_i^{\mathrm{valid}}$ is the number of valid timesteps up to the first success. 
This objective also complements KIR: keyframe-initialized rollouts tend to reach task resolution with fewer valid steps, and trajectory-length normalization increases their per-timestep contribution, so gradients are dominated by short, task-critical segments rather than long, drift-prone continuations.

\vspace{-1.0em}
\subsection{PACE: Policy–Aligned Co-Evolution}
\label{sec:pace}
\vspace{-0.6em}
While policy optimization proceeds entirely within the learned world model, the policy's action distribution continuously evolves and drifts away from the data used to train the initial world model. This inherent distribution shift leads to accumulating mismatch between the simulator and the improving policy, ultimately degrading the reliability of imagined rollouts.

To address this issue, we introduce \textbf{PACE}, a World Model--Policy co-evolution strategy. Instead of treating the world model as a fixed, static simulator throughout policy optimization, PACE allows the world model and VLA policy to evolve together throughout training.

Concretely, we realize this co-evolution through low-frequency, policy-driven refinement: we first train an initial world model, denoted as $\mathrm{WM}_{\mathrm{Base}}$, using trajectories collected from the base VLA policy. After the first stage of policy optimization within $\mathrm{WM}_{\mathrm{Base}}$, we collect a limited set of additional rollouts under the evolved policy and use them to further refine the world model. The refined model is referred to as $\mathrm{WM}_{\mathrm{Evo}}$. Importantly, this refinement is performed at very low frequency, distinguishing \emph{PACE} from classical model-based reinforcement learning methods, which continuously update the dynamics model at high frequency during policy optimization. 
This low-frequency refinement provides two key advantages. First, unlike real-world online RL, it does not require continuous human supervision or environment resets during policy training, significantly reducing operational overhead. 
Second, by aligning the world model with the evolving policy distribution, \emph{PACE} mitigates compounding model errors and maintains simulator reliability without sacrificing training stability.

\section{Experiments}

We conduct extensive experiments to evaluate the effectiveness of WoVR as a world-model-based reinforcement learning framework for post-training VLA policies.
Our experimental design aims to systematically answer the following three questions:

\begin{itemize}
    \item \textbf{Q1:} Is the proposed world model stable, controllable, and efficient enough to serve as a simulator for closed-loop reinforcement learning?
    \item \textbf{Q2:} Can WoVR effectively improve VLA task performance compared to existing world-model-based reinforcement learning methods?
    \item \textbf{Q3:} Do the policies optimized with WoVR reliably transfer to real-world manipulation tasks?
\end{itemize}

To answer these questions, we evaluate both the quality of the learned world model and the downstream policy performance.
For world model evaluation, we focus on long-horizon, action-conditioned video generation under closed-loop, chunk-by-chunk autoregressive inference.
We adopt standard perceptual and distributional metrics, including \textbf{LPIPS}~\citep{zhang2018unreasonable}, \textbf{FID}~\citep{heusel2017gans}, \textbf{FVD}~\citep{unterthiner2018towards} and \textbf{FloLPIPS}~\citep{danier2022flolpips} (Detailed calculations are provided in the Appendix~\ref{app:eval_metrics}.). 
We also report inference throughput (\textbf{FPS}) to quantify generation efficiency. For policy evaluation, we use task success rate (\textbf{SR}) as the primary metric.

We compare WoVR against several representative baselines spanning both \emph{world model quality} and \emph{policy optimization}. 
For world model quality, we include EVAC~\citep{jiang2025enerverseacenvisioningembodiedenvironments}, Cosmos-Predict2~\citep{nvidia_cosmos_predict2_2025} and OpenSora~\citep{opensora2}(the world-model backbone adopted in WMPO~\citep{zhu2025wmpoworldmodelbasedpolicy}). 
All compared models are evaluated under the same chunk-wise autoregressive generation protocol to ensure a fair comparison.
For policy optimization, we consider OpenVLA-OFT~\citep{kim2025fine}, a base VLA policy trained purely with imitation learning; GRPO (Online)~\citep{guo2025deepseek}, trained with real-environment interaction under the same rollout budget; and WMPO~\citep{zhu2025wmpoworldmodelbasedpolicy}, which performs reinforcement learning using OpenSora.

\subsection{Is the World Model Stable, Controllable, and Efficient?}

\begin{wraptable}{r}{0.66\textwidth}
\vspace{-0.8em}
\centering
\caption{\textbf{Comparison of different world models.}
Rollout denotes the rollout horizon length.}
\label{tab:wm_quality_full}
\scriptsize
\setlength{\tabcolsep}{2.2pt}
\renewcommand{\arraystretch}{1.08}

\sisetup{
  table-number-alignment = center,
  round-mode = places,
  round-precision = 3
}

\newcommand{\methodsep}{\addlinespace[1pt]\specialrule{0.35pt}{1.5pt}{1.5pt}}

\begin{tabular*}{\linewidth}{@{\extracolsep{\fill}}lcc
S[table-format=1.3]
S[table-format=3.3]
S[table-format=3.3]
S[table-format=1.3]@{}}
\toprule
\textbf{Method} & \textbf{Rollout} & \textbf{FPS} $\uparrow$ &
{\textbf{LPIPS} $\downarrow$}~\citep{zhang2018unreasonable} &
{\textbf{FID} $\downarrow$}~\citep{heusel2017gans} &
{\textbf{FVD} $\downarrow$}~\citep{unterthiner2018towards} &
{\textbf{FloLPIPS} $\downarrow$}~\citep{danier2022flolpips} \\
\midrule

\multirow{3}{*}{EVAC~\citep{jiang2025enerverseacenvisioningembodiedenvironments}}
& 512 & \multirow{3}{*}{1.35}
& 0.146 & 46.528 & 345.818 & 0.205 \\
& 256 &
& 0.130 & 49.153 & 354.983 & 0.192 \\
& 128 &
& 0.106 & 44.337 & 423.132 & 0.166 \\
\methodsep

\multirow{3}{*}{\makecell[l]{Cosmos-\\Predict2~\citep{nvidia_cosmos_predict2_2025}}}
& 512 & \multirow{3}{*}{3.50}
& 0.315 & 165.862 & 275.737 & 0.265 \\
& 256 &
& 0.226 & 106.324 & 203.853 & 0.306 \\
& 128 &
& 0.164 & 77.555 & 304.456 & 0.281 \\
\methodsep

\multirow{3}{*}{OpenSora~\citep{zhu2025wmpoworldmodelbasedpolicy}}
& 512 & \multirow{3}{*}{7.00}
& 0.105 & 38.478 & 89.391 & 0.156 \\
& 256 &
& 0.082 & 33.577 & 94.998 & 0.122 \\
& 128 &
& 0.069 & 33.413 & 111.643 & 0.113 \\
\methodsep

\multirow{3}{*}{WoVR (Ours)}
& 512 & \multirow{3}{*}{\bfseries 23.0}
& \bfseries \textbf{0.091} & \bfseries \textbf{34.252} & \bfseries \textbf{68.011} & \bfseries \textbf{0.154} \\
& 256 &
& \bfseries \textbf{0.063} & \bfseries \textbf{24.378} & \bfseries \textbf{50.041} & \bfseries \textbf{0.102} \\
& 128 &
& \bfseries \textbf{0.047} & \bfseries \textbf{18.553} & \bfseries \textbf{39.047} & \bfseries \textbf{0.079} \\

\bottomrule
\end{tabular*}
\vspace{-0.8em}
\end{wraptable}

\paragraph{Experimental Setup.}
We conduct all world model evaluations in the LIBERO environment~\citep{liu2023liberobenchmarkingknowledgetransfer}.
A total of 3{,}000 VLA rollout trajectories, each with a length of 512 frames, are collected to train the world models.
In addition, 200 held-out trajectories of the same length are used exclusively for evaluation.
We compare WoVR against three representative action-conditioned world models: EVAC, Cosmos-Predict2, and OpenSora as adopted in WMPO.

\paragraph{Quantitative Results.}
Table~\ref{tab:wm_quality_full} shows that WoVR outperforms all baselines across all metrics, indicating higher visual fidelity, stronger temporal consistency, and more accurate dynamics.
These improvements become more pronounced as the rollout horizon increases, suggesting that WoVR is more robust in long-horizon autoregressive generation.

Despite adopting a larger backbone (Wan, $\sim$5B) than OpenSora ($\sim$1.3B), WoVR achieves higher inference throughput by requiring only five diffusion steps and leveraging a 3D VAE for spatiotemporal latent encoding, whereas OpenSora typically relies on more sampling steps and a 2D VAE.

\subsection{Can WoVR Effectively Improve VLA Task Performance?}
\label{sec:q2}

\paragraph{Experimental Setup.}

\begin{wraptable}{r}{0.6\textwidth}
\vspace{-0.8em}
\centering
\caption{\textbf{Task success rates (\%) across LIBERO suites.}}
\label{tab:policy_performance}
\scriptsize
\small
\setlength{\tabcolsep}{3.5pt}
\renewcommand{\arraystretch}{1.04}
\begin{tabular*}{\linewidth}{@{\extracolsep{\fill}}lccccc@{}}
\toprule
\textbf{Method} & \textbf{Spatial} & \textbf{Object} & \textbf{Goal} & \textbf{Long} & \textbf{Avg} \\
\midrule

\rowcolor{blue!10}
\multicolumn{6}{c}{\textbf{One-Trajectory SFT}} \\
OpenVLA-OFT~\citep{kim2025fine}
& 63.6 & 36.4 & 48.2 & 13.8 & 40.5 \\
w/ GRPO (online)~\citep{guo2025deepseek}
& 66.6 
& 45.2 
& 52.2 
& 14.6 
& 44.6 \\
w/ WMPO~\citep{zhu2025wmpoworldmodelbasedpolicy}
& 67.8 
& 65.4 
& 56.6 
& 13.8
& 50.9 \\
w/ \textbf{Ours}
& \textbf{84.2}
& \textbf{80.8}
& \textbf{77.4}
& \textbf{35.8}
& \textbf{69.5} \\
\rowcolor{blue!8}
$\Delta$ 
& \textcolor{red}{+20.6}
& \textcolor{red}{+44.4}
& \textcolor{red}{+29.2}
& \textcolor{red}{+22.0}
& \textcolor{red}{+29.0} \\
\midrule

\rowcolor{blue!10}
\multicolumn{6}{c}{\textbf{Full-Trajectory SFT}} \\
OpenVLA-OFT~\citep{kim2025fine}
& 93.6 & 83.0 & 90.0 & 85.6 & 88.1 \\
w/ GRPO (online)~\citep{guo2025deepseek}
& 94.6 
& 86.2 
& 92.2 
& 85.8 
& 89.7 \\
w/ WMPO~\citep{zhu2025wmpoworldmodelbasedpolicy}
& 95.0 
& 94.8 
& 92.8 
& 87.0 
& 92.4 \\
w/ \textbf{Ours}
& \textbf{98.8}
& \textbf{98.8}
& \textbf{94.8}
& \textbf{91.4}
& \textbf{96.0} \\
\rowcolor{blue!8}
$\Delta$ 
& \textcolor{red}{+5.2}
& \textcolor{red}{+15.8}
& \textcolor{red}{+4.8}
& \textcolor{red}{+5.8}
& \textcolor{red}{+7.9} \\
\bottomrule
\end{tabular*}
\end{wraptable}

We conduct policy optimization experiments on LIBERO Spatial, Object, Goal, and Long suites~\citep{liu2023liberobenchmarkingknowledgetransfer}, each containing 10 tasks. 
Following SimpleVLA-RL~\citep{li2025simplevla}, we initialize from OpenVLA-OFT and consider two supervised fine-tuning settings: \emph{one-trajectory SFT} and \emph{full-trajectory SFT}. 

For a fair comparison, all methods use the same budget of 2,500 real-environment trajectories per suite. 
\textbf{GRPO} directly uses them for online policy optimization, while \textbf{WMPO} uses them to train a world model and then performs RL entirely in imagination. \textbf{WoVR} follows the two-stage policy--simulator co-evolution protocol in Sec.~\ref{sec:pace}: it first trains an initial world model with 1,500 trajectories collected by the base policy, optimizes the policy in imagination, and then refines the world model with 1,000 additional trajectories collected by the evolved policy. 

\paragraph{Quantitative Results.}

Table~\ref{tab:policy_performance} reports \textbf{SR} across LIBERO suites under two SFT initializations. 
Under the \emph{one-trajectory SFT} setting, GRPO yields only a marginal improvement (4.1 \% $\uparrow$), indicating that limited real-environment interaction is insufficient for effective online RL. 
WMPO improves the average success rate to 50.9\%, but shows no gain on LIBERO-Long (0 \% $\uparrow$), suggesting that a less reliable world model cannot support long-horizon policy optimization. 
In contrast, WoVR achieves the best performance across all suites, increasing the average success rate to 69.5\%, with especially large gains on Object and Long. 
Under the stronger \emph{full-trajectory SFT} setting, where the base policy is already strong (88.1\%), all methods have less room for improvement: GRPO reaches 89.7\% and WMPO reaches 92.4\%, while WoVR still achieves the highest average success rate of 95.9\% and consistently outperforms WMPO. 

These results show that world-model-based RL can outperform limited-budget online RL, especially when the simulator is sufficiently stable and policy-aligned. Additional ablations on \emph{PACE} and \emph{KIR} are provided in Appendix~\ref{app:ablation:pace}.

\subsection{Do Policies Optimized with WoVR Reliably Transfer to the Real World?}
\label{sec:q3}

\paragraph{Experimental Setup}

\begin{wrapfigure}{R}{0.50\textwidth}
    \centering
    \captionsetup[subfigure]{font=small,skip=1pt}
    \begin{subfigure}{0.32\linewidth}
        \centering
        \includegraphics[width=\linewidth]{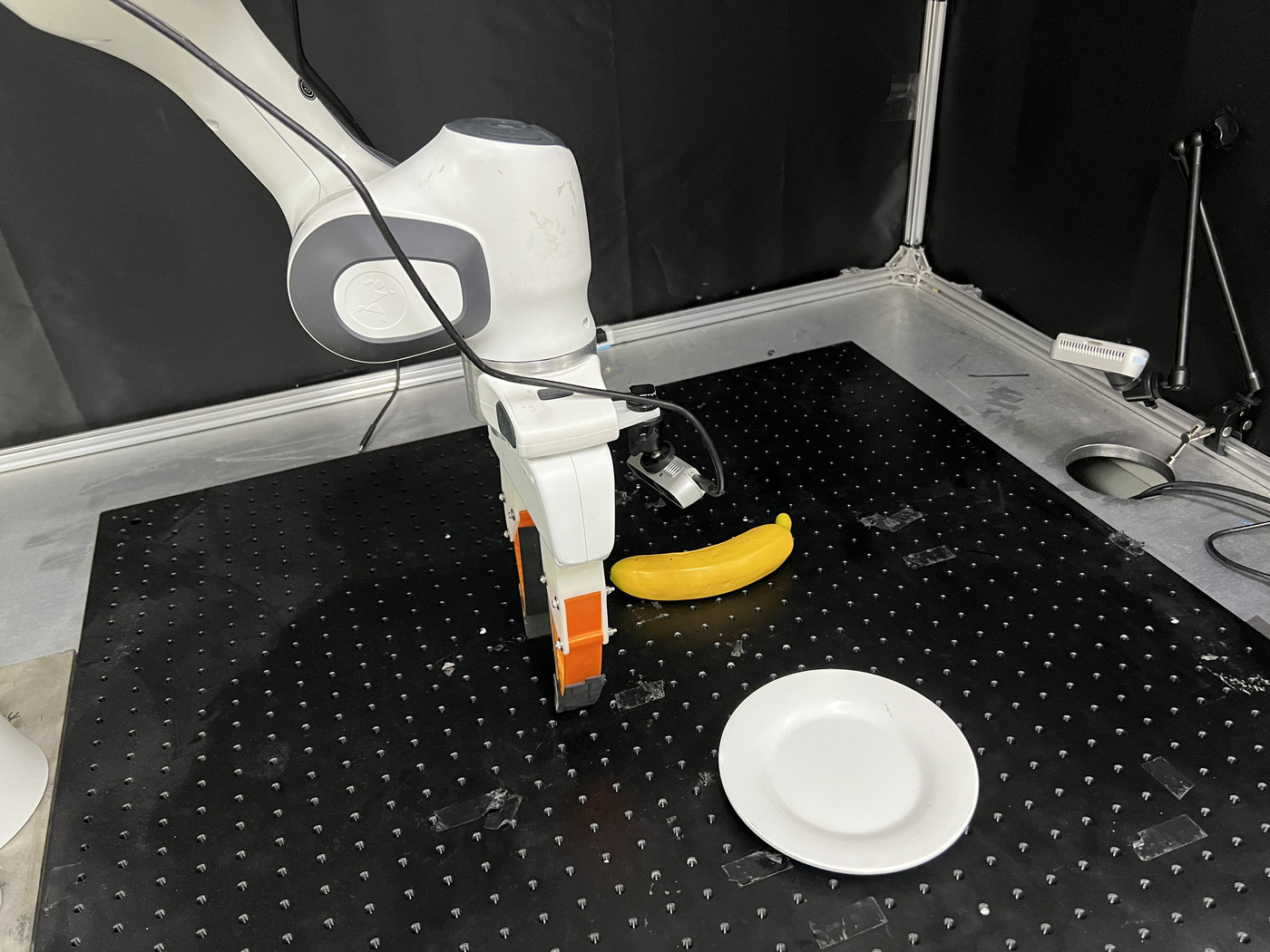}
        \caption{Pick banana}
    \end{subfigure}\hfill
    \begin{subfigure}{0.32\linewidth}
        \centering
        \includegraphics[width=\linewidth]{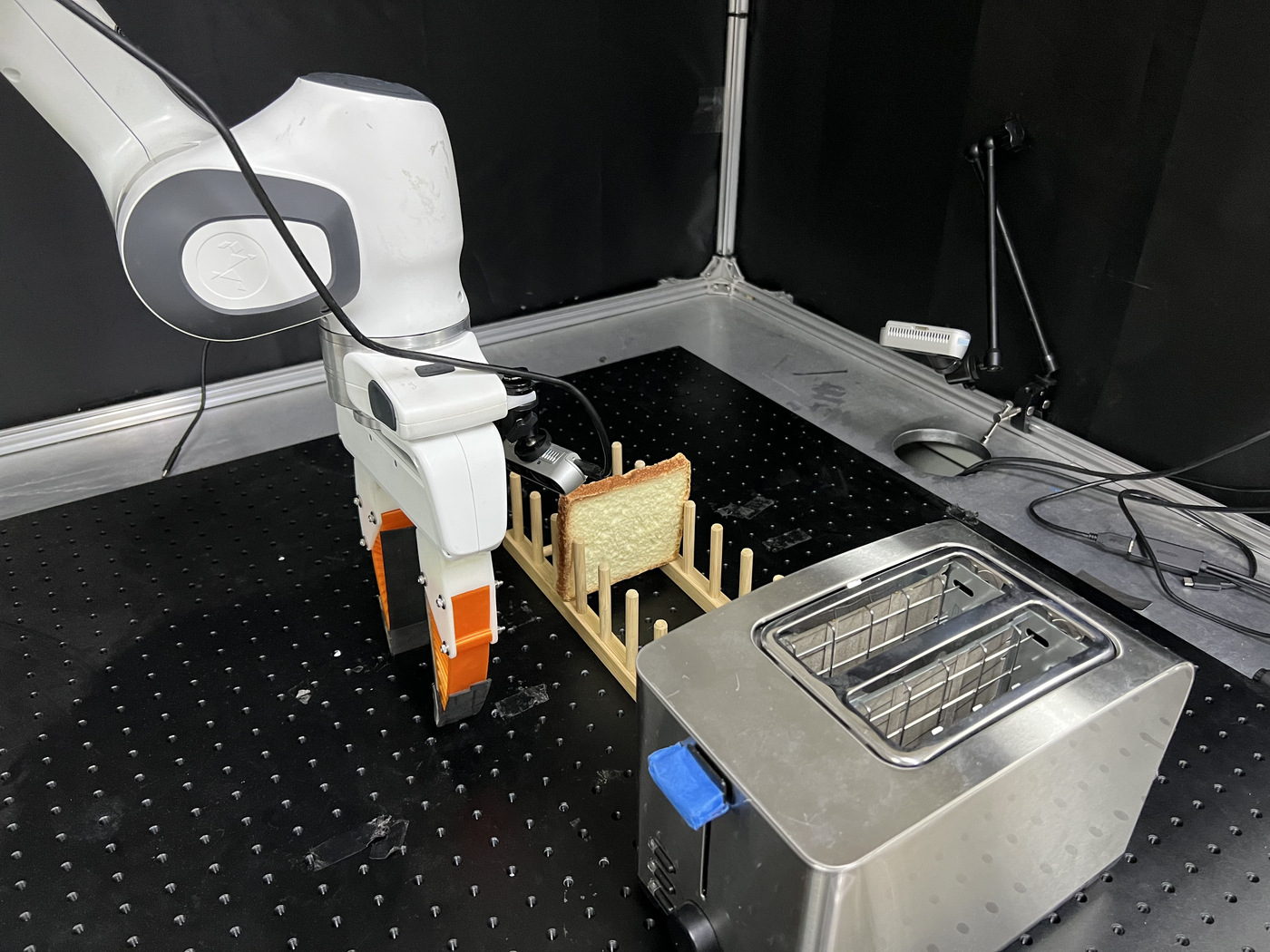}
        \caption{Pick bread}
    \end{subfigure}\hfill
    \begin{subfigure}{0.32\linewidth}
        \centering
        \includegraphics[width=\linewidth]{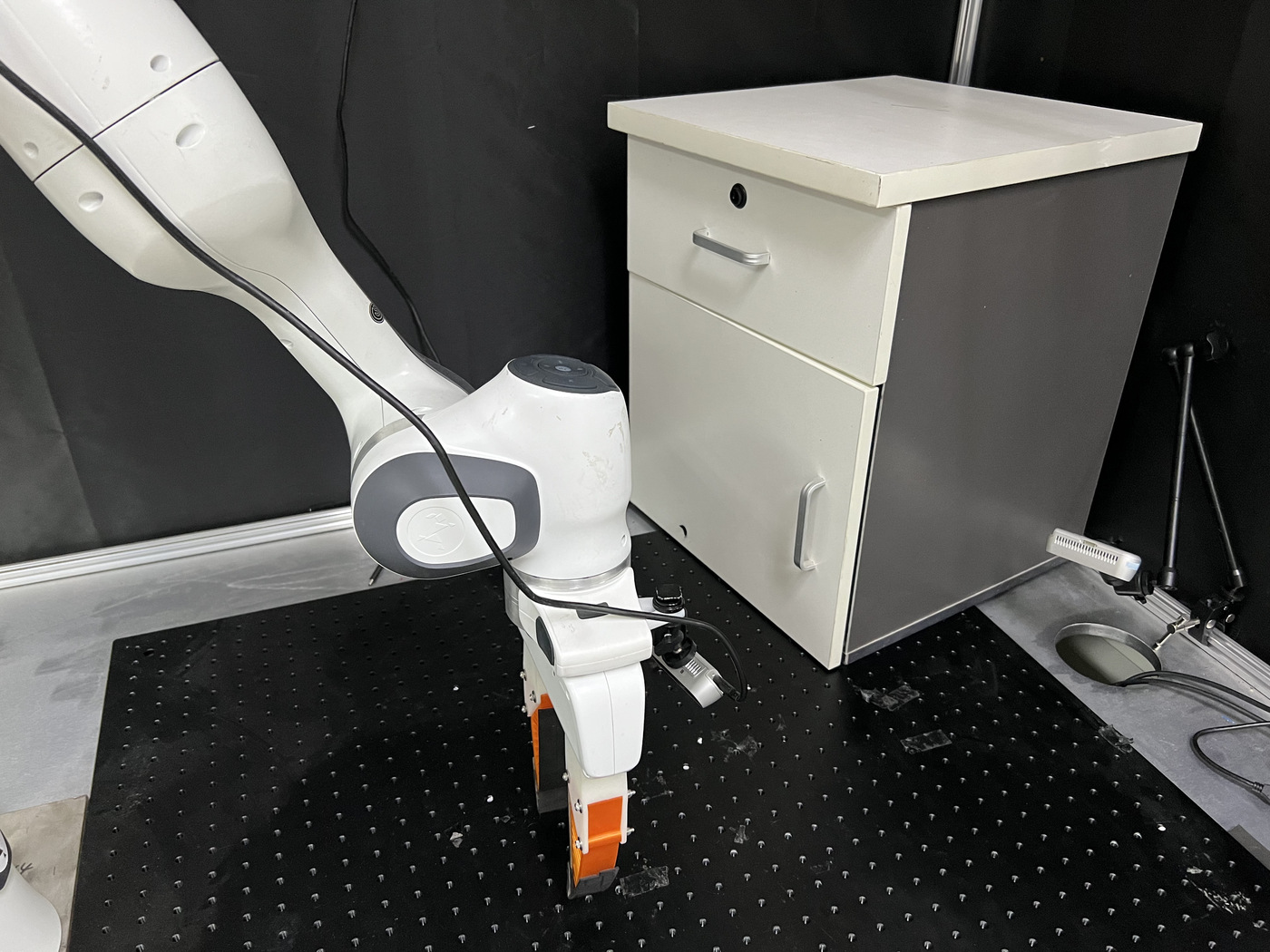}
        \caption{Open drawer}
    \end{subfigure}\\[0.2em]
    \begin{subfigure}{0.32\linewidth}
        \centering
        \includegraphics[width=\linewidth]{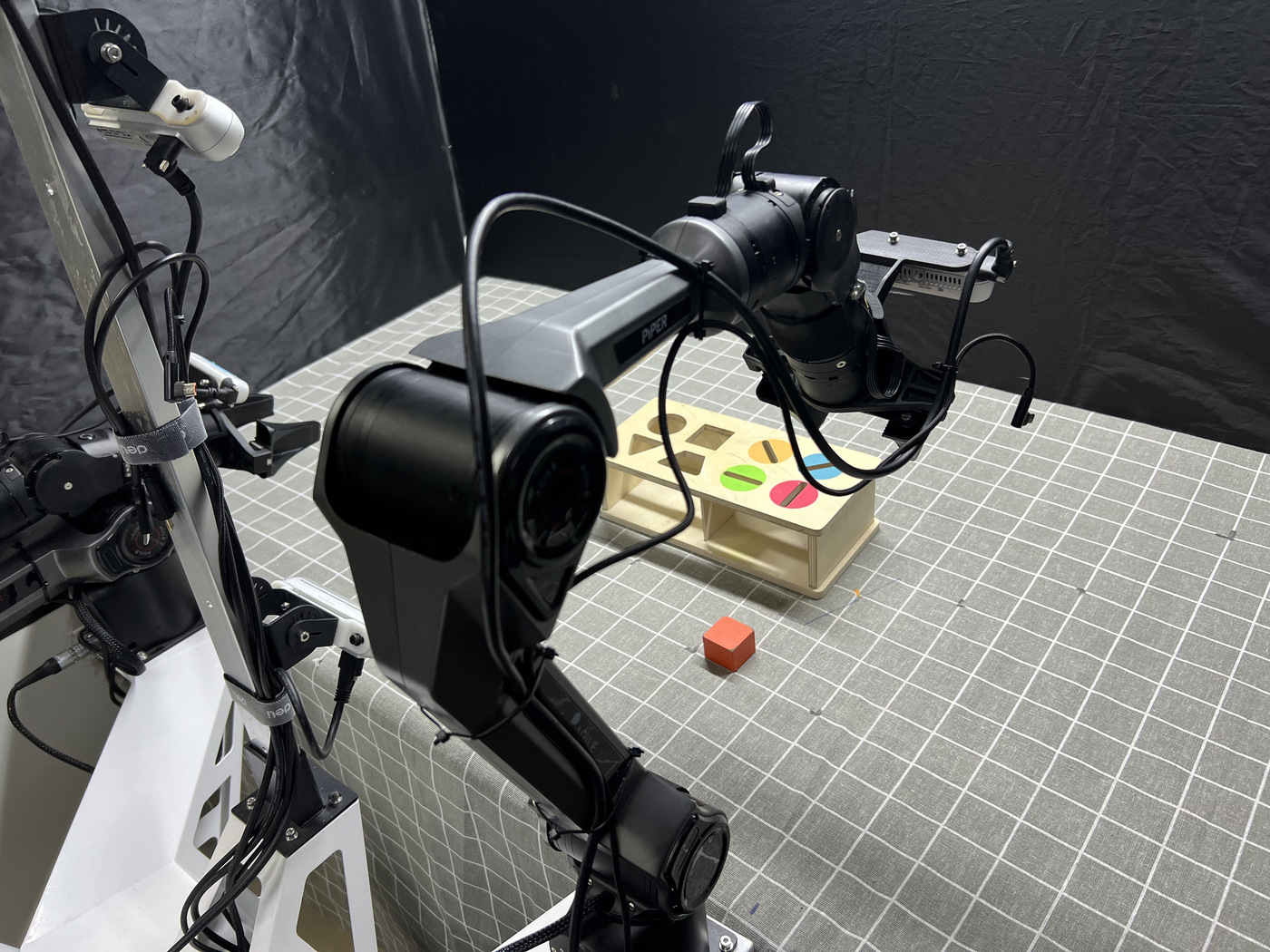}
        \caption{Pick cube}
    \end{subfigure}\hfill
    \begin{subfigure}{0.32\linewidth}
        \centering
        \includegraphics[width=\linewidth]{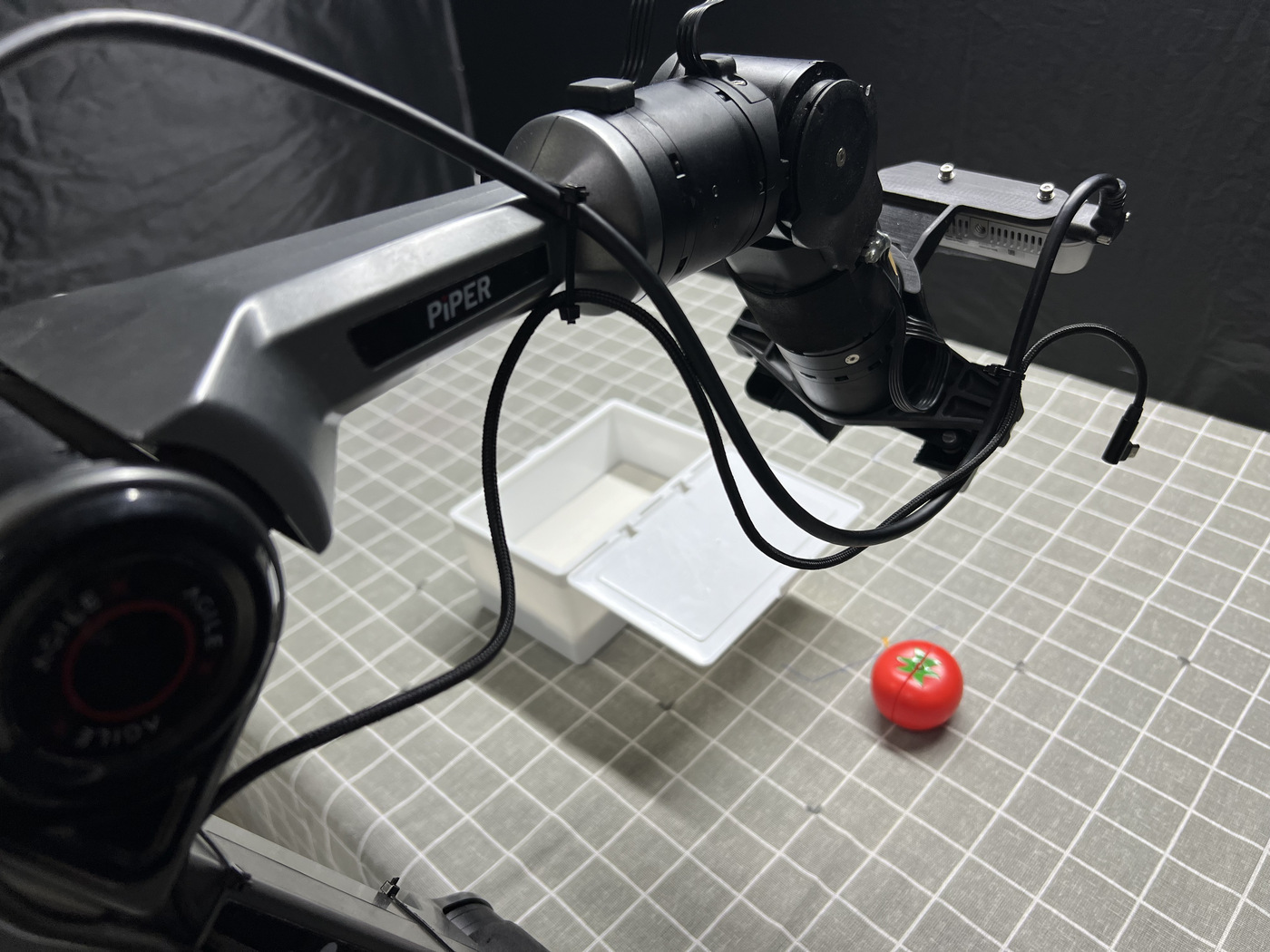}
        \caption{Pick tomato}
    \end{subfigure}\hfill
    \begin{subfigure}{0.32\linewidth}
        \centering
        \includegraphics[width=\linewidth]{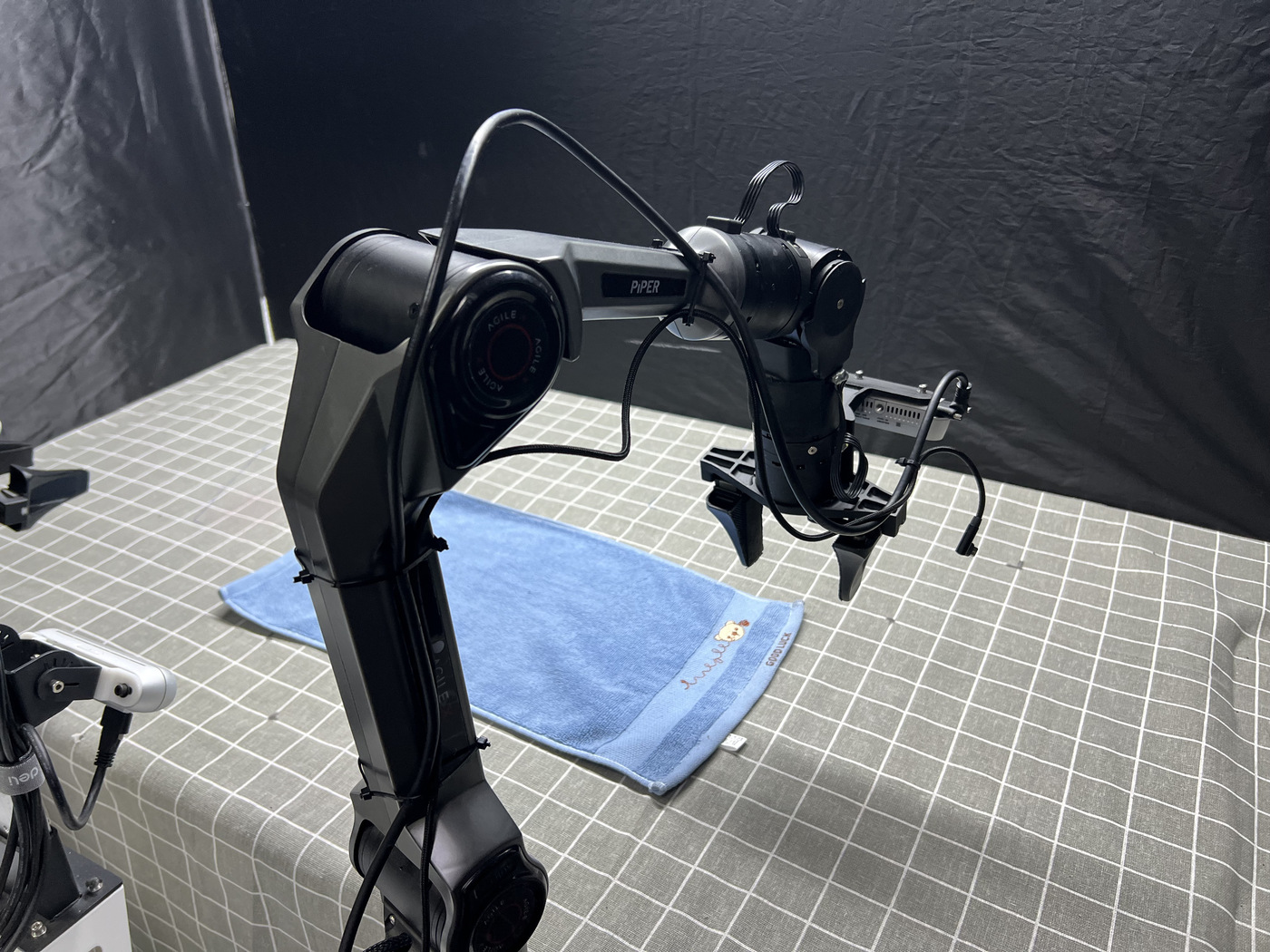}
        \caption{Fold towel}
    \end{subfigure}
    \caption{\textbf{Real-world experiments on two robotic platforms.}
    Top row: Franka Emika Panda; bottom row: AgileX Piper.}
    \label{fig:real_world_results}
\end{wrapfigure}

We conduct real-world transfer experiments on two robotic platforms with different control characteristics: the \textbf{Franka Emika Panda} and the \textbf{AgileX Piper}. 
As shown in Fig~\ref{fig:real_world_results}, we evaluate three manipulation tasks on each platform. 
Detailed task requirements are provided in Appendix~\ref{app:rw:detailed}.

For each platform, we train a \emph{multi-task} VLA policy and a corresponding world model over the three tasks. After training with WoVR, we directly deploy the optimized policies on the physical robots and evaluate their \textbf{SR} in real world, where each task is evaluated over 30 independent trials. More details can be seen in Appendix~\ref{sec:real_world}.

\paragraph{Quantitative Results.}

\begin{table}[t]
\centering
\caption{\textbf{Real-world success rates (\%, 30 trials/task).}}
\label{tab:real_world}
\scriptsize
\setlength{\tabcolsep}{3.6pt}
\renewcommand{\arraystretch}{1.05}
\begin{tabular}{lcccc@{\hspace{10pt}}cccc}
\toprule
\multirow{2}{*}{\textbf{Method}}
& \multicolumn{4}{c}{\textbf{Franka Emika Panda}}
& \multicolumn{4}{c}{\textbf{AgileX Piper}} \\
\cmidrule(lr){2-5}\cmidrule(lr){6-9}
& {Pick Banana} & {Pick Bread} & {Open Drawer} & \textbf{Avg}
& {Fold Towel} & {Pick Cube} & {Pick Tomato} & \textbf{Avg} \\
\midrule
OpenVLA-OFT~\citep{kim2025fine}
& 36.7 & 70.0 & 46.7 & 51.1
& 10.0 & 23.3 & 13.3 & 15.5 \\
w/ WMPO~\citep{zhu2025wmpoworldmodelbasedpolicy}
& 56.7 & 76.7 & 60.0 & 64.5
& 10.0 & 26.7 & 20.0 & 18.9 \\
w/ \textbf{Ours}
& \textbf{86.7} & \textbf{90.0} & \textbf{63.3} & \textbf{80.0}
& \textbf{20.0} & \textbf{33.3} & \textbf{33.3} & \textbf{28.9} \\
\rowcolor{blue!8}
$\Delta$
& \textcolor{red}{+50.0} & \textcolor{red}{+20.0} & \textcolor{red}{+16.6} & \textcolor{red}{+28.9}
& \textcolor{red}{+10.0} & \textcolor{red}{+10.0} & \textcolor{red}{+20.0} & \textcolor{red}{+13.4} \\
\bottomrule
\end{tabular}
\end{table}

Table~\ref{tab:real_world} reports real-world success rates on two robotic platforms. WMPO improves over the base policy, especially on Franka (+13.4 points), indicating that world-model-based policy optimization can provide useful learning signals. However, its gains are limited on the noisier AgileX Piper platform (+3.4 points), suggesting that naïve imagined RL remains vulnerable to world-model hallucination. In contrast, WoVR consistently achieves the best performance, improving the base policy by +28.9 points on Franka and +13.4 points on AgileX Piper. These results show that WoVR enables more stable and effective policy optimization in imagination, with strong transfer to both high-precision and noisier real-world robotic systems, without additional online interaction.

\paragraph{Generalization to Different VLA Backbones.}
To further evaluate the generality of WoVR, we conduct additional experiments using \textbf{$\pi_{0.5}$} as the VLA backbone on the AgileX Piper platform. 
We follow $\pi_{\mathrm{RL}}$~\citep{chen2026pitextttrlonlinerlfinetuning} and optimize the policy with Flow-SDE under the unchanged WoVR framework.

\begin{wraptable}{r}{0.53\textwidth}
\centering
\captionsetup{font=small}
\caption{\textbf{Generalization to $\pi_{0.5}$ on AgileX Piper.} Success rates (\%), 30 trials/task.}
\label{tab:real_world_openpi}
\small
\setlength{\tabcolsep}{5pt}
\renewcommand{\arraystretch}{1.04}
\begin{tabular}{@{}lcccc@{}}
\toprule
\textbf{Method} & \makecell{\textbf{Fold}\\\textbf{Towel}} & \makecell{\textbf{Pick}\\\textbf{Cube}} & \makecell{\textbf{Pick}\\\textbf{Tomato}} & \textbf{Avg} \\
\midrule
$\pi_{0.5}$-base
& 20.0 & 60.0 & 23.3 & 34.4 \\
w/ \textbf{Ours}
& \textbf{30.0}
& \textbf{86.7}
& \textbf{53.3}
& \textbf{56.7} \\
\rowcolor{blue!8}
$\Delta$
& \textcolor{red}{+10.0}
& \textcolor{red}{+26.7}
& \textcolor{red}{+30.0}
& \textcolor{red}{+22.3} \\
\bottomrule
\end{tabular}
\end{wraptable}

As shown in Table~\ref{tab:real_world_openpi}, WoVR consistently improves the $\pi_{0.5}$ policy across all evaluated tasks, raising the average success rate by +22.3 points.
The post-trained policy achieves high success rates on pick-and-place tasks.
Although Fold Towel is a more challenging deformable-object task, WoVR still brings a +10.0 point gain. 
These results demonstrate that the benefit of WoVR is not tied to a specific VLA architecture. 

\vspace{-1.0em}
\section{Conclusion}
\vspace{-0.6em}

In this work, we first identify hallucination in closed-loop imagined interaction as a central reliability bottleneck in world-model-based RL for VLA policy optimization.
To address this, we propose WoVR, a hallucination-aware framework that improves simulator stability with an action-controllable video world model, reduces effective prediction depth through KIR, and maintains policy--simulator alignment via PACE. 
Experiments on LIBERO and real-world manipulation tasks show that WoVR enables effective policy optimization.
Together, these results suggest that hallucination-controlled world models can serve as practical RL simulators, enabling a closed-loop path where online deployment data continually refine the simulator and further improve VLA policies in imagination.

\vspace{-0.8em}
\section{Limitations}
\vspace{-0.5em}

WoVR identifies hallucination in closed-loop imagined rollouts as a key reliability bottleneck for world-model-based RL, but our analysis remains primarily empirical. 
In particular, we do not provide a formal characterization of how hallucination propagates into policy optimization, nor a regret bound for the policy learned in the imagined world model relative to the optimal policy in the real environment. 
In addition, while WoVR demonstrates promising results on real-world tabletop manipulation tasks, its capability in substantially longer-horizon tasks and mobile manipulation remains underexplored. 
These settings may amplify error accumulation through extended interaction horizons, changing viewpoints, larger state spaces, and more diverse contact dynamics. 
Scaling WoVR to such scenarios may require large-scale world-model pretraining on more diverse embodied data, as well as more sophisticated mechanisms for suppressing compounding errors.

\acknowledgments{This work was supported by the New Generation Artificial Intelligence-National Science and Technology Major Project under Grant 2025ZD0122902; the National Natural Science Foundation of China (NSFC) under Grants 62406159 and 62293545; Zhongguancun Academy under Grant C20250301; the Beijing High Innovation Plan with Grant No. 202604841445; the Beijing Natural Science Foundation--Xiaomi Innovation Joint Fund under Grant L253007; the Beijing Major Science and Technology Project under Contract No. Z251100008125023; the Beijing Academy of Artificial Intelligence (BAAI); and the Shenzhen Science and Technology Program under Grant AI2026016.}

\clearpage
\bibliography{example}  

@inproceedings{zang2025rlinfvlaunifiedefficientframework,
  title        = {{RLux-VLA: A Unified and Efficient Framework for Reinforcement Learning of Vision-Language-Action Models}},
  author       = {Hongzhi Zang and Mingjie Wei and Si Xu and Yongji Wu and Zhen Guo and Yuanqing Wang and Hao Lin and Peihong Wang and Hua Yuan and Yixian Zhang and Liangzhi Shi and Yuqing Xie and Zhexuan Xu and Zhihao Liu and Kang Chen and Wenhao Tang and Quanlu Zhang and Weinan Zhang and Chao Yu and Yu Wang},
  booktitle    = {Proceedings of Robotics: Science and Systems},
  year         = {2026},
  doi          = {10.15607/RSS.2026.XXII.089},
}

@inproceedings{zang2026rlinfuserunifiedextensiblerealworld,
  title        = {{USER: A Unified and Extensible System for Online Real-World Policy Learning in Embodied AI}},
  author       = {Hongzhi Zang and Shu'ang Yu and Hao Lin and Tianxing Zhou and Zefang Huang and Zhen Guo and Xin Xu and Yuze Sheng and Jiakai Zhou and Shizhe Zhang and Feng Gao and Wenhao Tang and Yufeng Yue and Quanlu Zhang and Xinlei Chen and Chao Yu and Yu Wang},
  booktitle    = {Proceedings of Robotics: Science and Systems},
  year         = {2026},
  doi          = {10.15607/RSS.2026.XXII.037},
}

@article{chen2026pitextttrlonlinerlfinetuning,
  title        = {{$\pi_\texttt{RL}$: Online RL Fine-tuning for Flow-based Vision-Language-Action Models}},
  author       = {Kang Chen and Zhihao Liu and Tonghe Zhang and Zhen Guo and Si Xu and Hao Lin and Hongzhi Zang and Xiang Li and Quanlu Zhang and Zhaofei Yu and Guoliang Fan and Tiejun Huang and Yu Wang and Chao Yu},
  journal      = {arXiv preprint arXiv:2510.25889},
  year         = {2025},
  eprint       = {2510.25889},
  archiveprefix = {arXiv},
  primaryclass = {cs.LG},
}

@inproceedings{liu2026rlbringvlageneralization,
  title        = {{What Can RL Bring to VLA Generalization? An Empirical Study}},
  author       = {Liu, Jijia and Gao, Feng and Wei, Bingwen and Chen, Xinlei and Liao, Qingmin and Wu, Yi and Yu, Chao and Wang, Yu},
  booktitle    = {Advances in Neural Information Processing Systems},
  year         = {2025},
  volume       = {38, Main Conference},
  pages        = {97121--97151},
  publisher    = {Curran Associates, Inc.},
  doi          = {10.52202/085713-3248},
}

@inproceedings{chen2025conrft, 
    title={ConRFT: A Reinforced Fine-tuning Method for VLA Models via Consistency Policy}, 
    author={Yuhui Chen and Shuai Tian and Shugao Liu and Yingting Zhou and Haoran Li and Dongbin Zhao}, 
    booktitle={Proceedings of Robotics: Science and Systems, {RSS} 2025, Los Angeles, CA, USA, Jun 21-25, 2025}, 
    doi={10.15607/RSS.2025.XXI.019},
    year={2025},
}

@inproceedings{liu2026lifelongrft,
  title     = {Towards Long-Lived Robots: Continual Learning {VLA} Models via Reinforcement Fine-Tuning},
  author    = {Liu, Yuan and Li, Haoran and Tian, Shuai and Qin, Yuxing
               and Chen, Yuhui and Zheng, Yupeng and Huang, Yongzhen
               and Zhao, Dongbin},
  booktitle = {Proceedings of Robotics: Science and Systems, {RSS} 2026, Sydney, Australia, Jul, 2026},
  year      = {2026},
  doi       = {10.15607/RSS.2026.XXII.086},
}

@article{liu2026deg,
  title         = {Dual-Granularity Contrastive Reward via Generated Episodic Guidance for Efficient Embodied {RL}},
  author        = {Liu, Xin and Li, Yixuan and Chen, Yuhui and Qin, Yuxing
                   and Li, Haoran and Zhao, Dongbin},
  journal       = {arXiv preprint arXiv:2602.12636},
  year          = {2026},
  eprint        = {2602.12636},
  archivePrefix = {arXiv},
  primaryClass  = {cs.LG},
  doi           = {10.48550/arXiv.2602.12636},
}

@article{schulman2017proximal,
  title={Proximal policy optimization algorithms},
  author={Schulman, John and Wolski, Filip and Dhariwal, Prafulla and Radford, Alec and Klimov, Oleg},
  journal={arXiv preprint arXiv:1707.06347},
  year={2017}
}

@article{shao2024deepseekmathpushinglimitsmathematical,
      title={DeepSeekMath: Pushing the Limits of Mathematical Reasoning in Open Language Models},
      author={Zhihong Shao and Peiyi Wang and Qihao Zhu and Runxin Xu and Junxiao Song and Xiao Bi and Haowei Zhang and Mingchuan Zhang and Y. K. Li and Y. Wu and Daya Guo},
      year={2024},
      journal={arXiv preprint arXiv:2402.03300}
}

@article{guo2025deepseek,
  title        = {{DeepSeek-R1 incentivizes reasoning in LLMs through reinforcement learning}},
  author       = {Guo, Daya and Yang, Dejian and Zhang, Haowei and Song, Junxiao and Wang, Peiyi and Zhu, Qihao and Xu, Runxin and Zhang, Ruoyu and Ma, Shirong and Bi, Xiao and others},
  journal      = {Nature},
  year         = {2025},
  volume       = {645},
  number       = {8081},
  pages        = {633--638},
  doi          = {10.1038/s41586-025-09422-z},
}

@inproceedings{quevedo2025worldgymworldmodelenvironment,
  title        = {{WorldGym: World Model as An Environment for Policy Evaluation}},
  author       = {Quevedo, Julian and Sharma, Ansh Kumar and Sun, Yixiang and Suryavanshi, Varad and Liang, Percy and Yang, Sherry},
  booktitle    = {International Conference on Learning Representations},
  year         = {2026},
  pages        = {78932--78957},
}

@inproceedings{guo2025ctrlworldcontrollablegenerativeworld,
  title        = {{Ctrl-World: A Controllable Generative World Model for Robot Manipulation}},
  author       = {Guo, Yanjiang and Shi, Lucy and Chen, Jianyu and Finn, Chelsea},
  booktitle    = {International Conference on Learning Representations},
  year         = {2026},
  pages        = {6121--6138},
}

@inproceedings{jiang2025enerverseacenvisioningembodiedenvironments,
  title        = {{EnerVerse-AC: Envisioning Embodied Environments with Action Condition}},
  author       = {Yuxin Jiang and Shengcong Chen and Siyuan Huang and Liliang Chen and Pengfei Zhou and Yue Liao and Xindong He and Chiming Liu and Hongsheng Li and Maoqing Yao and Guanghui Ren},
  booktitle    = {NeurIPS 2025 Workshop on Embodied World Models for Decision Making},
  year         = {2025},
}

@article{xiao2025worldenvleveragingworldmodel,
      title={World-Env: Leveraging World Model as a Virtual Environment for VLA Post-Training},
      author={Junjin Xiao and Yandan Yang and Xinyuan Chang and Ronghan Chen and Feng Xiong and Mu Xu and Wei-Shi Zheng and Qing Zhang},
      year={2025},
      journal={arXiv preprint arXiv:2509.24948}
}

@inproceedings{zhu2025wmpoworldmodelbasedpolicy,
  title        = {{WMPO: World Model-based Policy Optimization for Vision-Language-Action Models}},
  author       = {Zhu, Fangqi and Yan, Zhengyang and Hong, Zicong and Shou, Quanxin and Ma, Xiao and Guo, Song},
  booktitle    = {International Conference on Learning Representations},
  year         = {2026},
  pages        = {62486--62502},
}

@article{tang2025hunyuangamecraft2instructionfollowinginteractivegame,
      title={Hunyuan-GameCraft-2: Instruction-following Interactive Game World Model},
      author={Junshu Tang and Jiacheng Liu and Jiaqi Li and Longhuang Wu and Haoyu Yang and Penghao Zhao and Siruis Gong and Xiang Yuan and Shuai Shao and Qinglin Lu},
      year={2025},
      journal={arXiv preprint arXiv:2511.23429}
}

@inproceedings{chen2024diffusionforcingnexttokenprediction,
  title        = {{Diffusion Forcing: Next-token Prediction Meets Full-Sequence Diffusion}},
  author       = {Chen, Boyuan and Mart\'{\i} Mons\'{o}, Diego and Du, Yilun and Simchowitz, Max and Tedrake, Russ and Sitzmann, Vincent},
  booktitle    = {Advances in Neural Information Processing Systems},
  year         = {2024},
  volume       = {37},
  pages        = {24081--24125},
  publisher    = {Curran Associates, Inc.},
  doi          = {10.52202/079017-0759},
}

@inproceedings{shin2025motionstreamrealtimevideogeneration,
  title        = {{MotionStream: Real-Time Video Generation with Interactive Motion Controls}},
  author       = {Shin, Joonghyuk and Li, Zhengqi and Zhang, Richard and Zhu, Jun-Yan and Park, Jaesik and Shechtman, Eli and Huang, Xun},
  booktitle    = {International Conference on Learning Representations},
  year         = {2026},
  pages        = {7381--7406},
}

@inproceedings{yang2025longliverealtimeinteractivelong,
  title        = {{LongLive: Real-time Interactive Long Video Generation}},
  author       = {Yang, Shuai and Huang, Wei and Chu, Ruihang and Xiao, Yicheng and Zhao, Yuyang and Wang, Xianbang and Li, Muyang and Xie, Enze and Chen, Yingcong and Lu, Yao and Han, Song and Chen, Yukang},
  booktitle    = {International Conference on Learning Representations},
  year         = {2026},
  pages        = {90037--90059},
}

@inproceedings{esser2024scalingrectifiedflowtransformers,
  title        = {{Scaling Rectified Flow Transformers for High-Resolution Image Synthesis}},
  author       = {Esser, Patrick and Kulal, Sumith and Blattmann, Andreas and Entezari, Rahim and M\"{u}ller, Jonas and Saini, Harry and Levi, Yam and Lorenz, Dominik and Sauer, Axel and Boesel, Frederic and Podell, Dustin and Dockhorn, Tim and English, Zion and Rombach, Robin},
  booktitle    = {Proceedings of the 41st International Conference on Machine Learning},
  year         = {2024},
  volume       = {235},
  pages        = {12606--12633},
  series       = {Proceedings of Machine Learning Research},
  publisher    = {PMLR},
}

@article{pan2026sopscalableonlineposttraining,
      title={SOP: A Scalable Online Post-Training System for Vision-Language-Action Models},
      author={Mingjie Pan and Siyuan Feng and Qinglin Zhang and Xinchen Li and Jianheng Song and Chendi Qu and Yi Wang and Chuankang Li and Ziyu Xiong and Zhi Chen and Yi Liu and Jianlan Luo},
      year={2026},
      journal={arXiv preprint arXiv:2601.03044}
}

@article{luo2025precise,
  title        = {{Precise and dexterous robotic manipulation via human-in-the-loop reinforcement learning}},
  author       = {Luo, Jianlan and Xu, Charles and Wu, Jeffrey and Levine, Sergey},
  journal      = {Science Robotics},
  year         = {2025},
  volume       = {10},
  number       = {105},
  pages        = {eads5033},
  doi          = {10.1126/scirobotics.ads5033},
}

@inproceedings{liu2023liberobenchmarkingknowledgetransfer,
  title        = {{LIBERO: Benchmarking Knowledge Transfer for Lifelong Robot Learning}},
  author       = {Liu, Bo and Zhu, Yifeng and Gao, Chongkai and Feng, Yihao and Liu, Qiang and Zhu, Yuke and Stone, Peter},
  booktitle    = {Advances in Neural Information Processing Systems},
  year         = {2023},
  volume       = {36},
  pages        = {44776--44791},
  publisher    = {Curran Associates, Inc.},
  doi          = {10.52202/075280-1939},
}

@inproceedings{kim2025fine,
  title        = {{Fine-Tuning Vision-Language-Action Models: Optimizing Speed and Success}},
  author       = {Moo Jin Kim and Chelsea Finn and Percy Liang},
  booktitle    = {Proceedings of Robotics: Science and Systems},
  year         = {2025},
  doi          = {10.15607/RSS.2025.XXI.017},
}

@inproceedings{black2024pi0visionlanguageactionflowmodel,
  title        = {{$\pi_0$: A Vision-Language-Action Flow Model for General Robot Control}},
  author       = {Kevin Black and Noah Brown and Danny Driess and Adnan Esmail and Michael Robert Equi and Chelsea Finn and Niccolo Fusai and Lachy Groom and Karol Hausman and Brian Ichter and others},
  booktitle    = {Proceedings of Robotics: Science and Systems},
  year         = {2025},
  doi          = {10.15607/RSS.2025.XXI.010},
}

@inproceedings{intelligence2025pi05visionlanguageactionmodelopenworld,
  title        = {{$\pi_{0.5}$: a Vision-Language-Action Model with Open-World Generalization}},
  author       = {Black, Kevin and Brown, Noah and Darpinian, James and Dhabalia, Karan and Driess, Danny and Esmail, Adnan and Equi, Michael Robert and Finn, Chelsea and Fusai, Niccolo and Galliker, Manuel Y. and Ghosh, Dibya and Groom, Lachy and Hausman, Karol and Ichter, Brian and Jakubczak, Szymon and Jones, Tim and others},
  booktitle    = {Proceedings of The 9th Conference on Robot Learning},
  year         = {2025},
  volume       = {305},
  pages        = {17--40},
  series       = {Proceedings of Machine Learning Research},
  publisher    = {PMLR},
}

@inproceedings{heusel2017gans,
  title        = {{GANs Trained by a Two Time-Scale Update Rule Converge to a Local Nash Equilibrium}},
  author       = {Heusel, Martin and Ramsauer, Hubert and Unterthiner, Thomas and Nessler, Bernhard and Hochreiter, Sepp},
  booktitle    = {Advances in Neural Information Processing Systems},
  year         = {2017},
  volume       = {30},
  publisher    = {Curran Associates, Inc.},
}

@article{unterthiner2018towards,
  title={Towards accurate generative models of video: A new metric \& challenges},
  author={Unterthiner, Thomas and Van Steenkiste, Sjoerd and Kurach, Karol and Marinier, Raphael and Michalski, Marcin and Gelly, Sylvain},
  journal={arXiv preprint arXiv:1812.01717},
  year={2018}
}

@inproceedings{zhang2018unreasonable,
  title={The unreasonable effectiveness of deep features as a perceptual metric},
  author={Zhang, Richard and Isola, Phillip and Efros, Alexei A and Shechtman, Eli and Wang, Oliver},
  booktitle={Proceedings of the IEEE conference on computer vision and pattern recognition},
  pages={586--595},
  year={2018}
}

@inproceedings{danier2022flolpips,
  title        = {{FloLPIPS: A Bespoke Video Quality Metric for Frame Interpolation}},
  author       = {Danier, Duolikun and Zhang, Fan and Bull, David},
  booktitle    = {2022 Picture Coding Symposium (PCS)},
  year         = {2022},
  pages        = {283--287},
  doi          = {10.1109/pcs56426.2022.10018062},
}

@article{opensora,
  title={Open-sora: Democratizing efficient video production for all},
  author={Zheng, Zangwei and Peng, Xiangyu and Yang, Tianji and Shen, Chenhui and Li, Shenggui and Liu, Hongxin and Zhou, Yukun and Li, Tianyi and You, Yang},
  journal={arXiv preprint arXiv:2412.20404},
  year={2024}
}

@article{opensora2,
  title        = {{Open-Sora 2.0: Training a Commercial-Level Video Generation Model in \$200k}},
  author       = {Zangwei Zheng and Xiangyu Peng and Yuxuan Lou and Chenhui Shen and Tom Young and Xinying Guo and Binluo Wang and Hang Xu and Hongxin Liu and Mingyan Jiang and others},
  journal      = {arXiv preprint arXiv:2503.09642},
  year         = {2025},
}

@misc{nvidia_cosmos_predict2_2025,
  author = {Pennington, Joel and Joshi, Pranjali and Bhide, Asawaree},
  title = {Develop Custom Physical AI Foundation Models with NVIDIA Cosmos Predict-2},
  year = {2025},
  month = {June 11},
  note = {NVIDIA Developer Blog}
}

@inproceedings{cui2024gapartmanip,
  title        = {{GAPartManip: A Large-Scale Part-Centric Dataset for Material-Agnostic Articulated Object Manipulation}},
  author       = {Cui, Wenbo and Zhao, Chengyang and Wei, Songlin and Zhang, Jiazhao and Geng, Haoran and Chen, Yaran and Li, Haoran and Wang, He},
  booktitle    = {2025 IEEE International Conference on Robotics and Automation (ICRA)},
  year         = {2025},
  pages        = {14791--14798},
  doi          = {10.1109/icra55743.2025.11128643},
}

@article{jiang2025world4rldiffusionworldmodels,
  title        = {{World4RL: Diffusion World Models for Policy Refinement With Reinforcement Learning for Robotic Manipulation}},
  author       = {Jiang, Zhennan and Liu, Kai and Qin, Yuxing and Tian, Shuai and Zheng, Yupeng and Zhou, Mingcai and Zhang, Zhengtao and Yu, Chao and Li, Haoran and Zhao, Dongbin},
  journal      = {IEEE Robotics and Automation Letters},
  year         = {2026},
  volume       = {11},
  number       = {10},
  pages        = {12240--12247},
  doi          = {10.1109/lra.2026.3728345},
}

@article{wan2025wanopenadvancedlargescale,
      title={Wan: Open and Advanced Large-Scale Video Generative Models},
      author={Team Wan and Ang Wang and Baole Ai and Bin Wen and Chaojie Mao and Chen-Wei Xie and Di Chen and Feiwu Yu and Haiming Zhao and Jianxiao Yang and others},
      year={2025},
      journal={arXiv preprint arXiv:2503.20314}
}

@article{chen2026tevir,
  title   = {{TeViR}: Text-to-Video Reward With Diffusion Models for Efficient Reinforcement Learning},
  author  = {Chen, Yuhui and Li, Haoran and Jiang, Zhennan
             and Wen, Haowei and Zhao, Dongbin},
  journal = {IEEE Transactions on Systems, Man, and Cybernetics: Systems},
  year    = {2026},
  volume  = {56},
  number  = {2},
  pages   = {893--905},
  month   = feb,
  doi     = {10.1109/TSMC.2025.3638818}
}

@inproceedings{liu2025fetchbot,
  title     = {{FetchBot}: Learning Generalizable Object Fetching in Cluttered Scenes via Zero-Shot {Sim2Real}},
  author    = {Liu, Weiheng and Wan, Yuxuan and Wang, Jilong
               and Kuang, Yuxuan and Cui, Wenbo and Shi, Xuesong
               and Li, Haoran and Zhao, Dongbin and Zhang, Zhizheng
               and Wang, He},
  booktitle = {Proceedings of The 9th Conference on Robot Learning},
  year      = {2025},
  editor    = {Lim, Joseph and Song, Shuran and Park, Hae-Won},
  series    = {Proceedings of Machine Learning Research},
  volume    = {305},
  pages     = {859--884},
  publisher = {PMLR},
}

@article{li2025vlarftvisionlanguageactionreinforcementfinetuning,
      title={VLA-RFT: Vision-Language-Action Reinforcement Fine-tuning with Verified Rewards in World Simulators},
      author={Hengtao Li and Pengxiang Ding and Runze Suo and Yihao Wang and Zirui Ge and Dongyuan Zang and Kexian Yu and Mingyang Sun and Hongyin Zhang and Donglin Wang and Weihua Su},
      year={2025},
      journal={arXiv preprint arXiv:2510.00406}
}

@article{li2025surveyvisionlanguageactionmodelsembodied,
  title   = {Survey of Vision-Language-Action Models for Embodied Manipulation},
  author  = {Li, Haoran and Chen, Yuhui and Cui, Wenbo and Liu, Weiheng
             and Liu, Kai and Zhou, Mingcai and Zhang, Zhengtao
             and Zhao, Dongbin},
  journal = {Acta Automatica Sinica},
  year    = {2026},
  volume  = {52},
  number  = {1},
  pages   = {18--51},
  doi     = {10.16383/j.aas.c250394},
}

@inproceedings{li2025simplevla,
  title        = {{SimpleVLA-RL: Scaling VLA Training via Reinforcement Learning}},
  author       = {Li, Haozhan and Zuo, Yuxin and Yu, Jiale and Zhang, Yuhao and Yang, Zhaohui and Zhang, Kaiyan and Zhu, Xuekai and Zhang, Yuchen and Chen, Tianxing and Cui, Ganqu and others},
  booktitle    = {International Conference on Learning Representations},
  year         = {2026},
  pages        = {125846--125869},
}

@inproceedings{lu2025vlarlmasterfulgeneralrobotic,
  title        = {{VLA-RL: Towards Masterful and General Robotic Manipulation with Scalable Reinforcement Learning}},
  author       = {Guanxing Lu and Wenkai Guo and Chubin Zhang and Yuheng Zhou and Haonan Jiang and Zifeng Gao and Yansong Tang and Ziwei Wang},
  booktitle    = {2026 IEEE/RSJ International Conference on Intelligent Robots and Systems (IROS)},
  year         = {2026},
}

@inproceedings{yuan2024policydecoratormodelagnosticonline,
  title        = {{Policy Decorator: Model-Agnostic Online Refinement for Large Policy Model}},
  author       = {Yuan, Xiu and Mu, Tongzhou and Tao, Stone and Fang, Yunhao and Zhang, Zhang and Su, Hao},
  booktitle    = {International Conference on Learning Representations},
  year         = {2025},
  pages        = {28129--28164},
}

@article{lei2025rl100performantroboticmanipulation,
  title        = {{Performant robotic manipulation with real-world reinforcement learning}},
  author       = {Lei, Kun and Li, Huanyu and Yu, Dongjie and Wei, Zhenyu and Guo, Lingxiao and Jiang, Zhennan and Wang, Ziyu and Liang, Shiyu and Xu, Huazhe},
  journal      = {Science Robotics},
  year         = {2026},
  volume       = {11},
  number       = {116},
  pages        = {eaed6267},
  doi          = {10.1126/scirobotics.aed6267},
}

@article{li2025grrlgoingdexterousprecise,
      title={GR-RL: Going Dexterous and Precise for Long-Horizon Robotic Manipulation}, 
      author={Yunfei Li and Xiao Ma and Jiafeng Xu and Yu Cui and Zhongren Cui and Zhigang Han and Liqun Huang and Tao Kong and Yuxiao Liu and Hao Niu and Wanli Peng and Jingchao Qiao and Zeyu Ren and Haixin Shi and Zhi Su and Jiawen Tian and Yuyang Xiao and Shenyu Zhang and Liwei Zheng and Hang Li and Yonghui Wu},
      year={2025},
      journal={arXiv preprint arXiv:2512.01801}
}

@inproceedings{zheng2025xvlasoftpromptedtransformerscalable,
  title        = {{X-VLA: Soft-Prompted Transformer as Scalable Cross-Embodiment Vision-Language-Action Model}},
  author       = {Zheng, Jinliang and Li, Jianxiong and Wang, Zhihao and Liu, Dongxiu and Kang, Xirui and Feng, Yuchun and Zheng, Yinan and Zou, Jiayin and Chen, Yilun and Zeng, Jia and Wang, Tai and Zhang, Ya-Qin and Liu, Jingjing and Zhan, Xianyuan},
  booktitle    = {International Conference on Learning Representations},
  year         = {2026},
  pages        = {60580--60606},
}

@inproceedings{zhou2020continuityrotationrepresentationsneural,
  title        = {{On the Continuity of Rotation Representations in Neural Networks}},
  author       = {Zhou, Yi and Barnes, Connelly and Lu, Jingwan and Yang, Jimei and Li, Hao},
  booktitle    = {Proceedings of the IEEE/CVF Conference on Computer Vision and Pattern Recognition (CVPR)},
  year         = {2019},
  pages        = {5745--5753},
}

@article{Qwen3-VL,
      title={Qwen3-VL Technical Report}, 
      author={Shuai Bai and Yuxuan Cai and Ruizhe Chen and others},
	  journal={arXiv preprint arXiv:2511.21631},
      year={2025}
}

@article{guo2026vlawiterativecoimprovementvisionlanguageaction,
      title={VLAW: Iterative Co-Improvement of Vision-Language-Action Policy and World Model}, 
      author={Yanjiang Guo and Tony Lee and Lucy Xiaoyang Shi and Jianyu Chen and Percy Liang and Chelsea Finn},
      year={2026},
      journal={arXiv preprint arXiv:2602.12063},
}

@inproceedings{yang2026riseselfimprovingrobotpolicy,
  title        = {{Self-Improving Robot Policy with Compositional World Model}},
  author       = {Jiazhi Yang and Kunyang Lin and Wencong Zhang and Jinwei Li and Tianwei Lin and Longyan Wu and Ya-Qin Zhang and Hao Zhao and Ping Luo and Zhizhong Su and Hongyang Li and Xiangyu Yue and Li Chen},
  booktitle    = {Proceedings of Robotics: Science and Systems},
  year         = {2026},
  doi          = {10.15607/RSS.2026.XXII.012},
}

@inproceedings{fei2025srposelfreferentialpolicyoptimization,
  title        = {{SRPO: Self-Referential Policy Optimization for Vision-Language-Action Models}},
  author       = {Fei, Senyu and Wang, Siyin and Ji, Li and Li, Ao and Zhang, Shiduo and Liu, Liming and Hou, Jinlong and Gong, Jingjing and Zhao, Xianzhong and Qiu, Xipeng},
  booktitle    = {Proceedings of the IEEE/CVF Conference on Computer Vision and Pattern Recognition (CVPR)},
  year         = {2026},
  pages        = {6718--6728},
}

@inproceedings{sun2026atomvlascalableposttrainingrobotic,
  title        = {{AtomVLA: Scalable Post-Training for Robotic Manipulation via Predictive Latent World Models}},
  author       = {Xiaoquan Sun and Zetian Xu and Chen Cao and Zonghe Liu and Yihan Sun and Jingrui Pang and Ruijian Zhang and Zhen Yang and Kang Pang and Dingxin He and Mingqi Yuan and Jiayu Chen},
  booktitle    = {2026 IEEE/RSJ International Conference on Intelligent Robots and Systems (IROS)},
  year         = {2026},
}

@article{hung2025nora15visionlanguageactionmodeltrained,
  title        = {{NORA-1.5: A Vision-Language-Action Model Trained using World Model- and Action-based Preference Rewards}},
  author       = {Chia-Yu Hung and Navonil Majumder and Haoyuan Deng and Liu Renhang and Yankang Ang and Amir Zadeh and Chuan Li and Dorien Herremans and Ziwei Wang and Soujanya Poria},
  journal      = {arXiv preprint arXiv:2511.14659},
  year         = {2025},
}

@article{zhang2025reinforcingactionpoliciesprophesying,
  title        = {{Reinforcing Action Policies by Prophesying}},
  author       = {Jiahui Zhang and Ze Huang and Chun Gu and Zipei Ma and Li Zhang},
  journal      = {arXiv preprint arXiv:2511.20633},
  year         = {2025},
}

@article{pei2026evolutionvla,
  title        = {{The evolution of end-to-end Vision-Language-Action (VLA) architectures in robotics}},
  author       = {Pei, Jingjing and Zheng, Xiaoyin and Liu, Yang and Zhu, Bike and Wang, Daifeng and An, Jiajun and Voyles, Richard and Ma, Xin},
  journal      = {Robot Learning},
  year         = {2026},
  volume       = {3},
  number       = {2},
  pages        = {0010},
  doi          = {10.55092/rl20260010},
}

@article{zhang2026conformalizedppo,
  title        = {{Conformalized proximal policy optimization: statistical uncertainty quantification for principled exploration in reinforcement learning}},
  author       = {Zhang, Bailing and Chen, Genlang and Feng, Weiguo},
  journal      = {Robot Learning},
  year         = {2026},
  volume       = {3},
  number       = {2},
  pages        = {0011},
  doi          = {10.55092/rl20260011},
}

\clearpage
\appendix

\section*{Appendix Contents}
\vspace{0.2em}
\noindent\rule{\linewidth}{0.5pt}
\vspace{0.9em}

\begingroup
\setlength{\parindent}{0pt}
\setlength{\parskip}{0.68em}
\renewcommand{\arraystretch}{1.08}

\noindent\hyperref[app:gpu_allocation]{\textbf{A\quad GPU Allocation Strategy}}\dotfill\pageref{app:gpu_allocation}\par

\vspace{0.35em}
\noindent\hyperref[app:impl]{\textbf{B\quad Implemental Details}}\dotfill\pageref{app:impl}\par
\hspace{2.2em}\hyperref[app:impl:world_model]{B.1\quad Implemental Details of world model}\dotfill\pageref{app:impl:world_model}\par
\hspace{2.2em}\hyperref[sec:rm]{B.2\quad Implemental Details of Reward Model}\dotfill\pageref{sec:rm}\par
\hspace{4.0em}\hyperref[app:rm:sparse]{B.2.1\quad Sparse Reward Model}\dotfill\pageref{app:rm:sparse}\par
\hspace{4.0em}\hyperref[app:rm:dense]{B.2.2\quad Dense Reward Modeling}\dotfill\pageref{app:rm:dense}\par
\hspace{2.2em}\hyperref[app:impl:kir]{B.3\quad Implemental Details of KIR}\dotfill\pageref{app:impl:kir}\par

\vspace{0.35em}
\noindent\hyperref[app:eval_metrics]{\textbf{C\quad Evaluation Metrics}}\dotfill\pageref{app:eval_metrics}\par

\vspace{0.35em}
\noindent\hyperref[app:ablation]{\textbf{D\quad Ablation Study}}\dotfill\pageref{app:ablation}\par
\hspace{2.2em}\hyperref[app:ablation:wm]{D.1\quad Ablation on World Model Mechanisms}\dotfill\pageref{app:ablation:wm}\par
\hspace{2.2em}\hyperref[app:ablation:pace]{D.2\quad Ablation on PACE and KIR}\dotfill\pageref{app:ablation:pace}\par
\hspace{2.2em}\hyperref[sec:ablation_reward]{D.3\quad Ablation on Reward Modeling}\dotfill\pageref{sec:ablation_reward}\par

\noindent\hyperref[sec:fail_mode]{\textbf{E\quad Qualitative Failure Mode Analysis}}\dotfill\pageref{sec:fail_mode}\par

\vspace{0.35em}
\noindent\hyperref[sec:real_world]{\textbf{F\quad Real-World Experiments}}\dotfill\pageref{sec:real_world}\par
\hspace{2.2em}\hyperref[app:rw:hardware]{F.1\quad Hardware Setup}\dotfill\pageref{app:rw:hardware}\par
\hspace{2.2em}\hyperref[app:rw:detailed]{F.2\quad Detailed Task Requirements}\dotfill\pageref{app:rw:detailed}\par
\hspace{2.2em}\hyperref[app:rw:practical]{F.3\quad Practical Considerations for Real-World Deployment}\dotfill\pageref{app:rw:practical}\par
\hspace{2.2em}\hyperref[app:rw:task]{F.4\quad Task Setting}\dotfill\pageref{app:rw:task}\par

\endgroup

\vfill
\noindent\rule{\linewidth}{0.5pt}
\vspace{0.4em}

\clearpage

\section{GPU Allocation Strategy}
\label{app:gpu_allocation}

The reinforcement-learning pipeline can be decomposed into three components: \textit{Generation}, \textit{Simulator}, and \textit{Training}.
In WoVR, the \textit{Simulator} is instantiated by the learned \emph{world model}, which generates the next observation given the current observation and action.
In the rollout phase, \textit{Generation} performs policy inference to produce an (optionally chunked) action from the current observation, while the \textit{Simulator} executes the action and returns the next observation; this closed-loop interaction repeats until a batch of trajectories is collected. In the optimization phase, \textit{Training} updates the VLA policy using the collected trajectories, after which the system alternates back to rollout for the next iteration.

\begin{figure}[h]
    \centering
    \includegraphics[width=0.6\linewidth]{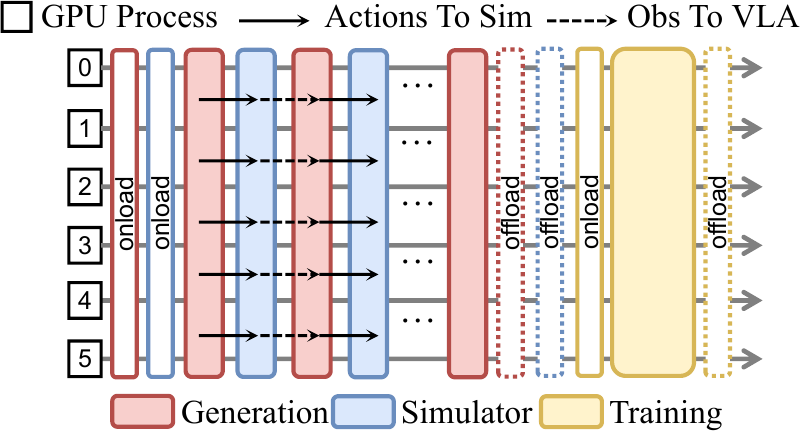}
    \caption{Collocated GPU Allocation Strategy}
    \label{fig:gpu}
\end{figure}

Following the system abstraction in RLinf-VLA, WoVR adopts a collocated (shared) GPU allocation strategy, where the three RL pipeline components co-exist on the same set of GPUs, with the \textit{Simulator} implemented as the world-model rollout module.
Unlike physical simulators that require dedicated device-side state, WoVR's simulator is a neural network; thus, offload/onload can be naturally realized by swapping only the \emph{model parameters} between GPU and host memory, without migrating any external simulator state.
In its original form, collocated execution relied on frequent GPU$\leftrightarrow$CPU offload/onload to keep only one component resident on GPUs at a time; however, in embodied settings the simulator and generator must interact iteratively, making per-interaction offload/onload prohibitively expensive.
Therefore, we use the modified collocated strategy: offload/onload for Generation and Simulator happens only at the beginning and end of the rollout phase, avoiding repeated transfers during closed-loop imagined interaction, as illustrated in Fig.~\ref{fig:gpu}.

\section{Implemental Details}
\label{app:impl}

\subsection{Implemental Details of world model}
\label{app:impl:world_model}

The world model generates imagined trajectories chunk by chunk in the latent space of the video diffusion backbone.
At rollout step $t$, the anchored visual context is formed by concatenating the episode's initial frame and the latest generated memory frames, i.e.,
$[o_0,\, o_{t-c:t}]$.
After VAE encoding, the corresponding latent context $[z_0,\, z_{t'-c':t'}]$ is concatenated with Gaussian noise latents for the next chunk.
The action-conditioned DiT blocks then denoise these latents under the dual-channel action conditioning shown in Fig.~\ref{fig:model_arch}, and the decoded future frames are appended to the rollout context.
Repeating this procedure yields long-horizon closed-loop imagined trajectories under policy actions.

We train the model with the Rectified Flow objective~\citep{esser2024scalingrectifiedflowtransformers}.
Let $x_1=z_{t'+1:t'+H'}$ denote the target future latents and $x_0\sim\mathcal{N}(0,\mathbf{I})$ denote Gaussian noise.
Given an interpolation time $\tau\in[0,1]$, the linearly interpolated state is constructed as:
\begin{equation}
x_{\tau} = (1-\tau)x_0 + \tau x_1 .
\end{equation}
The corresponding target velocity field along this interpolation trajectory is:
\begin{equation}
v_{\tau} = \frac{d x_{\tau}}{d \tau} = x_1 - x_0 .
\end{equation}
the model $u(\cdot;\phi)$ predicts the velocity $v_{\tau}$ from the intermediate latent $x_{\tau}$:
\begin{equation}
\mathcal{L} = \mathbb{E}_{x_0, x_1, c, \tau} \left[ \left\| u(x_{\tau}, c, \tau; \phi) - v_{\tau} \right\|^2 \right],
\label{eq:loss}
\end{equation}
where $c$ includes the anchored visual context and the action sequence.

To improve robustness under closed-loop inference, we inject diffusion noise into the non-reference context latents during training.
\begin{equation}
    z_{t'-c':t'} = \tau_{ctx}* z_{t'-c':t'} + (1-\tau_{ctx}) *  \epsilon,
\end{equation}
where $\tau_{ctx}$ is a value close to 1 and $\epsilon$ denotes noise.
The initial reference frame remains clean, while recent memory latents are mildly corrupted before being used as context.
This augmentation reduces the train--inference gap caused by conditioning on self-generated frames and mitigates brittle context copying during long-horizon autoregressive rollout~\citep{chen2024diffusionforcingnexttokenprediction,guo2025ctrlworldcontrollablegenerativeworld}.

\subsection{Implemental Details of Reward Model}
\label{sec:rm}

\subsubsection{Sparse Reward Model}
\label{app:rm:sparse}

For sparse supervision, we use a learned reward classifier that predicts task success from the predicted next observation. Concretely, given the generated observation $\tilde{o}_{t+1}$ from the world model, the sparse reward model $R_{\psi}^{\mathrm{sp}}$ estimates the probability of task success, and the binary reward is defined as
\begin{equation}
r^{\mathrm{sp}}_{t+1} = \mathbb{I}\left( R_{\psi}^{\mathrm{sp}}(\tilde{o}_{t+1}) \ge 0.5 \right),
\end{equation}
where $\mathbb{I}(\cdot)$ denotes the indicator function. Following HiL-SERL~\citep{luo2025precise}, the sparse reward classifier is implemented as a lightweight network and trained with binary cross-entropy loss on labeled success states, defined as

\begin{equation}
\label{eq:RCJFunc}
\begin{aligned}
\mathcal{L}_{binary}(\psi)
= - \frac{1}{N} \sum_{i=1}^N \Big[ r_i \log C_\psi(x_i) + (1-r_i)\log\big(1-C_\psi(x_i)\big)
\Big].
\end{aligned}
\end{equation}

\subsubsection{Dense Reward Modeling}
\label{app:rm:dense}

To provide finer-grained learning signals beyond binary success, we further introduce a dense reward model built on top of Qwen3-VL~\cite{Qwen3-VL}. Instead of directly regressing a scalar reward, we formulate dense reward prediction as an ordinal visual progress estimation problem. Given four consecutive observations
$\tilde{o}_{t-2:t+1} = [\tilde{o}_{t-2}, \tilde{o}_{t-1}, \tilde{o}_{t}, \tilde{o}_{t+1}]$
and the task description $l$, the dense reward model $R_{\psi}^{\mathrm{de}}$ predicts a discrete progress level
\begin{equation}
\ell_t \in \{0,1,\dots,10\},
\end{equation}
where $0$ denotes no progress and $10$ denotes task completion. The dense reward is then obtained by normalization:
\begin{equation}
r^{\mathrm{de}}_t = \frac{\ell_t}{10}.
\end{equation}

The model takes as input four consecutive RGB frames together with the task text, and outputs logits over 11 reward levels. Architecturally, we adopt a Qwen3-VL backbone with LoRA adaptation and attach a lightweight MLP reward head for classification. During training, labels are constructed automatically from trajectory rewards: failed trajectories are assigned level $0$, successful states are assigned level $10$, and pre-success states are linearly mapped to intermediate levels $1$--$9$ according to their temporal proximity to the first successful state. To mitigate class imbalance, we further rebalance samples across reward levels during dataset construction.

The dense reward model is trained with a standard cross-entropy objective:
\begin{equation}
\mathcal{L}_{\mathrm{dense}}(\psi)=
\mathbb{E}_{(\tilde{o}_{t-2:t+1},d,\ell_t)}
\left[
-\log p_{\psi}^{\mathrm{de}}(\ell_t \mid \tilde{o}_{t-2:t+1}, d)
\right].
\end{equation}

\subsection{Implemental Details of KIR}
\label{app:impl:kir}

Keyframe-Initialized Rollouts (KIR) are implemented as a change to the initial context fed into the world model, rather than as a change to the model architecture.
The Wan encoder used by our world model is a causal 3D VAE whose temporal input length follows the form $1+4m$.
In our rollout implementation, we use five frames (m=1) in total: one persistent reference frame plus four temporal context frames.

For standard long-horizon rollout from the episode start, the first frame is repeated to construct a valid VAE input:
\begin{equation}
\mathcal{C}^{\mathrm{init}}
= [o_0,o_0,o_0,o_0,o_0]
= \operatorname{Repeat}(o_0, 5),
\end{equation}
and the first generated chunk is
\begin{equation}
\tilde{o}_{1:9}
= W_{\phi}\!\left(\mathcal{C}^{\mathrm{init}}, a_{0:8}\right),
\end{equation}
where $W_{\phi}$ denotes the action-conditioned world model.
Equivalently, when comparing against a KIR rollout at a task-specific horizon, the non-KIR context can be written as
\begin{equation}
\tilde{o}_{1:9}
= W_{\phi}\!\left(\operatorname{Repeat}(o_0,5), a_{0:8}\right),
\end{equation}
which removes the local pre-failure context and relies only on the initial observation.
This initialization forces the world model to imagine the entire prefix before reaching task-critical states, which increases the effective depth of error accumulation.

KIR instead initializes imagined rollouts near a task-critical state.
Ideally, for a trajectory that is about to fail around time $T_{\mathrm{KIR}}+1$, we would use the first frame together with the last four frames before the failure transition:
\begin{equation}
\mathcal{C}^{\mathrm{KIR}}_{T}
= [o_0, o_{T_{\mathrm{KIR}}-3}, o_{T_{\mathrm{KIR}}-2}, o_{T_{\mathrm{KIR}}-1}, o_{T_{\mathrm{KIR}}}],
\end{equation}
and generate the first imagined chunk from this keyframe context:
\begin{equation}
\tilde{o}_{T_{\mathrm{KIR}+1}:T_{\mathrm{KIR}}+9}
= W_{\phi}\!\left(\mathcal{C}^{\mathrm{KIR}}_{T}, a_{T_{\mathrm{KIR}}:T_{\mathrm{KIR}}+8}\right).
\end{equation}
The first frame $o_0$ remains the global anchor, while $o_{T_{\mathrm{KIR}}-3:T_{\mathrm{KIR}}}$ provides the local pre-failure context.

In practice, we avoid manually cherry-picking keyframes for every trajectory.
For each task, we set a task-level keyframe index $T_{\mathrm{KIR}}$ based on when the base policy typically approaches a critical failure region.
This fixed task-level choice keeps KIR inexpensive and reproducible while still placing the first imagined chunk near the decisive part of the task.
After the first KIR chunk is generated, subsequent chunks follow the same autoregressive procedure as the standard world-model rollout, using the first frame anchor and the latest generated memory frames as context.

\section{Evaluation Metrics}
\label{app:eval_metrics}

We adopt LPIPS~\citep{zhang2018unreasonable}, FID~\citep{heusel2017gans}, FVD~\citep{unterthiner2018towards}, FloLPIPS~\citep{danier2022flolpips}, and FPS as evaluation metrics for generated videos. Specifically:
\begin{itemize}
    \item \textbf{LPIPS} (Learned Perceptual Image Patch Similarity) measures frame-level perceptual similarity using deep visual features;
\begin{equation}
d\left(x, x_{0}\right) = \sum_{l} \frac{1}{H_{l} W_{l}} \sum_{h,w} \left\| w_{l} \odot \left( \hat{F}^{l}_{h,w}(x) - \hat{F}^{l}_{h,w}(x_{0}) \right) \right\|_{2}^{2}
\end{equation}
where $x$ and $x_0$ denote the generated and real frames, $\hat{F}^{l}_{h,w}(\cdot)$ denotes the normalized deep feature at spatial location $(h,w)$ of layer $l$, and $w_l$ is the learned channel-wise weighting.

    \item \textbf{FID} (Fréchet Inception Distance) evaluates the distributional similarity between generated and real frames based on image-level feature statistics;
\begin{equation}
\mathrm{FID} = \left\| \mu_{r} - \mu_{g} \right\|_{2}^{2} + \mathrm{Tr}\!\left( \Sigma_{r} + \Sigma_{g} - 2 \left( \Sigma_{r} \Sigma_{g} \right)^{1/2} \right)
\end{equation}
where $(\mu_r,\Sigma_r)$ and $(\mu_g,\Sigma_g)$ denote the mean and covariance of image-level features extracted from real and generated frames, respectively.

    \item \textbf{FVD} (Fréchet Video Distance) extends the FID formulation from image-level features to video-level features. 
    It evaluates the distributional similarity between real and generated videos by computing the mean and covariance statistics over spatiotemporal video representations;
\begin{equation}
\mathrm{FVD} = \left\| \mu_{r}^{v} - \mu_{g}^{v} \right\|_{2}^{2} + 
\mathrm{Tr}\!\left( \Sigma_{r}^{v} + \Sigma_{g}^{v} - 2 \left( \Sigma_{r}^{v} \Sigma_{g}^{v} \right)^{1/2} \right)
\end{equation}
where $(\mu_r^v,\Sigma_r^v)$ and $(\mu_g^v,\Sigma_g^v)$ denote the mean and covariance of video-level features extracted from real and generated videos, respectively.

    \item \textbf{FloLPIPS} measures perceptual similarity after motion alignment along estimated optical-flow trajectories;
\begin{equation}
\mathrm{FloLPIPS}
=
\frac{1}{T-1}
\sum_{t=1}^{T-1}
\sum_l
\frac{
\sum_{h,w}
M_t(h,w)
\left\|
w_l \odot
\left(
\hat{F}^{l}_{h,w}(V_t)
-
\hat{F}^{l}_{h,w}(\hat{V}_t)
\right)
\right\|_2^2
}{
\sum_{h,w} M_t(h,w)
}
\end{equation}
The motion weight is defined as:
\begin{equation}
M_t(h,w)
=
\left\|
\mathrm{Flow}(V_t, V_{t+1})_{h,w}
-
\mathrm{Flow}(\hat{V}_t, \hat{V}_{t+1})_{h,w}
\right\|_2
\end{equation}

where $\hat{V}_{t}$ and $V_{t}$ denote the generated and real frames at time $t$. 
$\hat{F}^{l}_{h,w}(\cdot)$ denotes the normalized feature extracted by the perceptual network at the $l$-th layer and spatial location $(h,w)$, and $w_l$ is the learned perceptual weight for that layer. 
$\mathrm{Flow}(V_t,V_{t+1})$ and $\mathrm{Flow}(\hat{V}_t,\hat{V}_{t+1})$ denote the optical flow between two adjacent frames in the real and generated videos, respectively. 
$M_t(h,w)$ measures the discrepancy between the motion pattern in the real video and that in the generated video at location $(h,w)$ based on optical flow. 

    \item \textbf{FPS} measures the number of frames generated per second, quantifying the generation efficiency of the world model.
\end{itemize}

\section{Ablation Study}
\label{app:ablation}

\subsection{Ablation on World Model Mechanisms}
\label{app:ablation:wm}

We first conduct ablation studies on the core design choices of the proposed world model, aiming to understand how different context modeling mechanisms affect long-horizon video generation stability.
Specifically, we investigate the following factors:
(i) the number of memory frames used as visual context,
(ii) the use of a fixed reference frame, and
(iii) the effect of adding noise to context frames during training.

\paragraph{Experimental Variants.}
We compare the full WoVR model against three ablated variants:
\begin{itemize}
    \item WoVR w/o ref, which removes the fixed reference frame from the context window;
    \item WoVR w. mem=1, which uses only a single-frame context;
    \item WoVR w/o noisy context, which disables noise injection on context frames during training.
\end{itemize}

All variants are trained and evaluated on the \textbf{LIBERO-Spatial} suite only. We train the world model using 1{,}500 VLA rollout trajectories and evaluate on a held-out set of 24 trajectories.

\paragraph{Quantitative Results.}

Table~\ref{tab:ablation_world_model} reports the quantitative results measured by LPIPS, FID, FVD and FloLPIPS under different rollout horizons. Compared to using a single-frame context, employing a multi-frame context with a fixed reference anchor significantly improves performance across all metrics.

\begin{table*}[ht]
\centering
\caption{\textbf{Ablation study on world model mechanisms (LIBERO-Spatial).}
Rollout denotes the rollout horizon length.}
\label{tab:ablation_world_model}
\setlength{\tabcolsep}{6pt}
\renewcommand{\arraystretch}{1.15}

\sisetup{
  table-number-alignment = center,
  round-mode = places,
  round-precision = 3
}

\begin{tabular}{
l c
S[table-format=1.3]
S[table-format=2.3]
S[table-format=3.3]
S[table-format=1.3]
}
\toprule
& & \multicolumn{4}{c}{\textbf{Metrics}} \\
\cmidrule(lr){3-6}
\textbf{Method} & \textbf{Rollout} &
\textbf{LPIPS} $\downarrow$ &
\textbf{FID} $\downarrow$ &
\textbf{FVD} $\downarrow$ &
\textbf{FloLPIPS} $\downarrow$ \\
\midrule

\multirow{3}{*}{WoVR (Ours)}
& 512 & \bfseries 0.091 & \bfseries 36.687 & \bfseries 73.493 & \bfseries 0.154 \\
& 256 & \bfseries 0.069 & \bfseries 27.238 & \bfseries 63.948 & \bfseries 0.110 \\
& 128 & \bfseries 0.051 & \bfseries 20.780 & \bfseries 49.017 & \bfseries 0.081 \\
\addlinespace[2pt]

\multirow{3}{*}{WoVR w/o ref}
& 512 & 0.133 & 73.942 & 123.502 & 0.168 \\
& 256 & 0.089 & 49.406 & 86.000 & 0.116 \\
& 128 & 0.064 & 35.559 & 86.146 & 0.090 \\

\multirow{3}{*}{WoVR w. mem=1}
& 512 & 0.120 & 64.501 & 86.042 & 0.165 \\
& 256 & 0.086 & 46.790 & 81.742 & 0.117 \\
& 128 & 0.065 & 36.047 & 79.605 & 0.095 \\
\addlinespace[2pt]

\multirow{3}{*}{WoVR w/o noisy context}
& 512 & 0.099 & 44.712 & 77.284 & 0.160 \\
& 256 & 0.074 & 31.691 & 61.660 & 0.115 \\
& 128 & 0.054 & 23.444 & 58.836 & 0.085 \\
\addlinespace[2pt]

\bottomrule
\end{tabular}
\end{table*}

\begin{figure}[ht]
    \centering
    \includegraphics[width=0.9\linewidth]{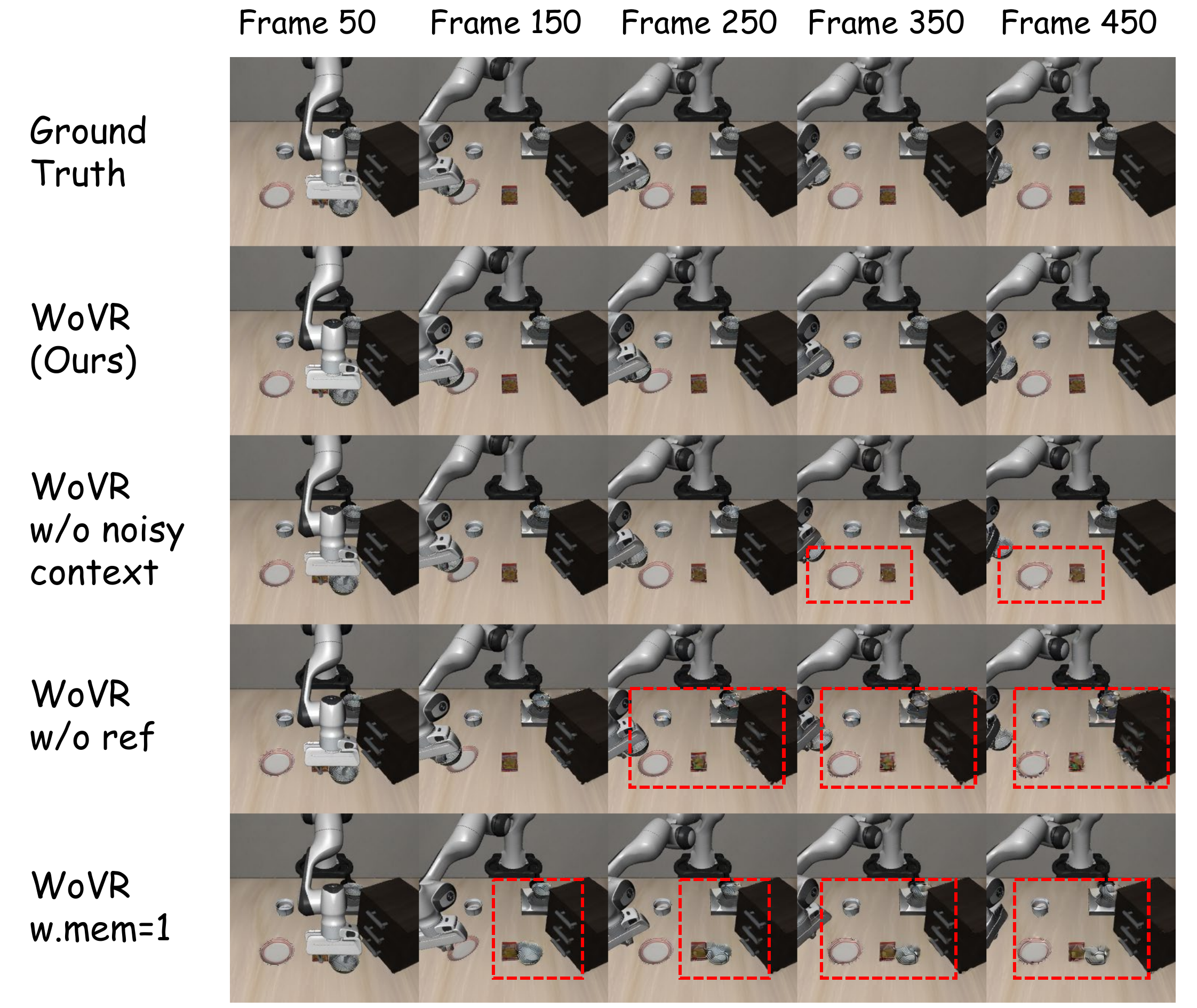}
    \caption{\textbf{Qualitative ablation results on LIBERO-Spatial.}
Ablated variants exhibit error accumulation and visual drift under long-horizon rollouts, while the full WoVR model remains stable and consistent with the ground truth.}
    \label{fig:ablation_qualitative}
\end{figure}

To better understand the failure modes behind these quantitative trends, we provide qualitative comparisons in Fig.~\ref{fig:ablation_qualitative}.
As shown in the figure, models without a fixed reference frame or noisy context exhibit noticeable spatial drift and object disappearance over long-horizon rollouts, whereas the full WoVR model remains visually stable and consistent with the ground truth.

Removing the reference frame leads to a clear degradation in performance, especially under longer rollout horizons.
This result suggests that anchoring the context with a fixed reference frame effectively suppresses error accumulation in the autoregressive feedback loop, which is critical for maintaining stability in long-horizon video generation.

Furthermore, disabling noise injection on context frames also results in noticeable performance drops.
While the degradation is moderate for short rollouts, the gap becomes more pronounced as the rollout length increases.
This observation indicates that adding mild noise to context frames improves robustness in long-horizon generation by reducing over-reliance on precise conditioning inputs, thereby alleviating the train--inference gap.

Overall, these results demonstrate that the proposed context modeling strategy---combining a fixed reference frame, a multi-frame memory window, and noisy context augmentation---plays a crucial role in stabilizing long-horizon video generation.
Together, these mechanisms enable WoVR to maintain high fidelity and temporal consistency under closed-loop autoregressive inference, providing a more reliable simulator for downstream reinforcement learning.

\subsection{Ablation on PACE and KIR}
\label{app:ablation:pace}

We further ablate two interaction-level reliability mechanisms in WoVR: PACE, which aligns the world model with the updated policy distribution, and KIR, which initializes imagined rollouts near task-critical states to reduce effective prediction depth.
This ablation is designed to separate the benefit of collecting policy-aligned data from the benefit of keyframe-initialized imagined interaction.

\paragraph{Experimental Setup.}
We compare four WoVR variants under the one-trajectory SFT policy-optimization protocol from Sec.~\ref{sec:q2}:
\begin{itemize}
    \item \textbf{WoVR w/o PACE (1{,}500 base)}, which trains the world model only on 1{,}500 trajectories rolled out by the base policy;
    \item \textbf{WoVR w/o PACE (2{,}500 base)}, which trains the world model on 2{,}500 trajectories rolled out by the base policy, increasing the amount of data without changing the policy distribution;
    \item \textbf{WoVR w/o KIR}, which keeps the PACE data protocol but removes keyframe initialization, so imagined rollouts are initialized from the episode start;
\end{itemize}

\begin{table}[ht]
\centering
\caption{\textbf{Ablation study on PACE and KIR.}
All variants optimize the policy through imagined rollouts. Avg is computed over the two reported suites.}
\label{tab:ablation_pace}
\small
\setlength{\tabcolsep}{3.2pt}
\renewcommand{\arraystretch}{1.04}
\begin{tabular*}{\linewidth}{@{\extracolsep{\fill}}lccc@{}}
\toprule
\textbf{Method} & \textbf{Spatial} & \textbf{Object} & \textbf{Avg} $\uparrow$ \\
\midrule
\rowcolor{blue!10}
\multicolumn{4}{c}{\textbf{One-Trajectory SFT}} \\
OpenVLA-OFT-base~\citep{kim2025fine}
& 63.6 & 36.4 & 50.0 \\
WoVR w/o PACE (1{,}500 base)
& 75.4 & 76.2 & 75.8 \\
WoVR w/o PACE (2{,}500 base)
& 77.8 & 77.2 & 77.5 \\
WoVR w/o KIR
& 81.6 & 77.8 & 79.7 \\
\textbf{WoVR} (1{,}500 base + 1{,}000 aligned)
& \textbf{84.2} & \textbf{80.8} & \textbf{82.5} \\
\rowcolor{blue!8}
$\Delta$
& \textcolor{red}{+20.6}
& \textcolor{red}{+44.4}
& \textcolor{red}{+32.5} \\
\bottomrule
\end{tabular*}
\end{table}

\paragraph{Results.}

Table~\ref{tab:ablation_pace} reports the completed ablations on LIBERO-Spatial and LIBERO-Object.
WoVR achieves the best performance on both suites, reaching an average success rate of 82.5\% over the two reported suites.
The two variants without PACE isolate the effect of additional base-policy data: increasing the world-model training set from 1{,}500 to 2{,}500 base-policy trajectories improves the two-suite average only from 75.8\% to 77.5\%.
This modest +1.7-point gain suggests that simply adding more data from the same base-policy distribution brings limited benefit once that distribution is sufficiently covered.

\begin{figure*}[t]
    \centering
    \includegraphics[width=\textwidth]{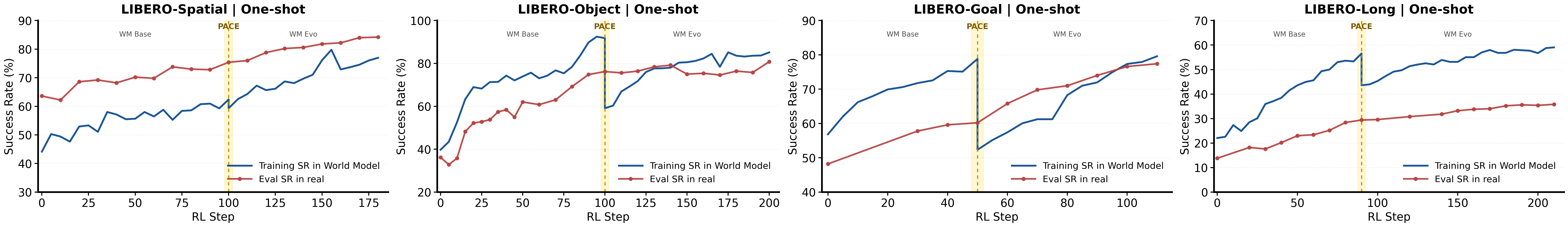}
    \caption{\textbf{One-trajectory SFT training and real-evaluation curves under WoVR.}
    Blue denotes the training success rate in the world model, red denotes the evaluation success rate in the real environment, and the shaded vertical band marks the PACE transition from $\mathrm{WM}_{\mathrm{Base}}$ to $\mathrm{WM}_{\mathrm{Evo}}$.}
    \label{fig:ablation_pace_kir_curves}
\end{figure*}

Fig.~\ref{fig:ablation_pace_kir_curves} shows how WoVR evolves during one-trajectory SFT training.
Before the PACE transition, the imagined training success rate can keep improving while the real-evaluation curve lags behind, revealing a growing mismatch between the evolving policy and the base world model.
After updating the simulator from $\mathrm{WM}_{\mathrm{Base}}$ to $\mathrm{WM}_{\mathrm{Evo}}$, the imagined training curve is re-calibrated to the policy-induced distribution and becomes more consistent with real-environment evaluation.
This post-PACE alignment indicates that rollout data collected from the evolved policy helps correct accumulated simulator mismatch and mitigates hallucination-induced over-optimism in imagined RL.
Together with the quantitative results in Table~\ref{tab:ablation_pace}, these curves support the role of PACE in making world-model training signals more faithful to real-world policy performance.

The KIR ablation further shows the importance of initializing imagined rollouts near task-critical states.
Removing KIR reduces performance from 84.2\% to 81.6\% on Spatial and from 80.8\% to 77.8\% on Object, lowering the two-suite average from 82.5\% to 79.7\%.
These results indicate that KIR improves imagined interaction even when PACE is retained, by shortening the effective prediction depth and reducing early error accumulation.
Together, the PACE and KIR ablations suggest that WoVR benefits from both policy-aligned simulator refinement and keyframe-initialized rollout design.

\subsection{Ablation on Reward Modeling}
\label{sec:ablation_reward}

We further analyze the impact of different reward modeling choices on policy optimization, comparing a lightweight ResNet-based sparse reward model with a Qwen3-VL-based dense reward model.

\paragraph{Experimental Setup.}
We keep all other components fixed and vary only the reward model used during policy optimization. The sparse reward model predicts binary task success, while the dense reward model provides fine-grained progress signals. Both variants are evaluated on the same training setup and metrics as in Sec.~\ref{sec:q2}.

\paragraph{Results and Analysis.}

\begin{figure}[ht]
    \centering
    \includegraphics[width=0.9\linewidth]{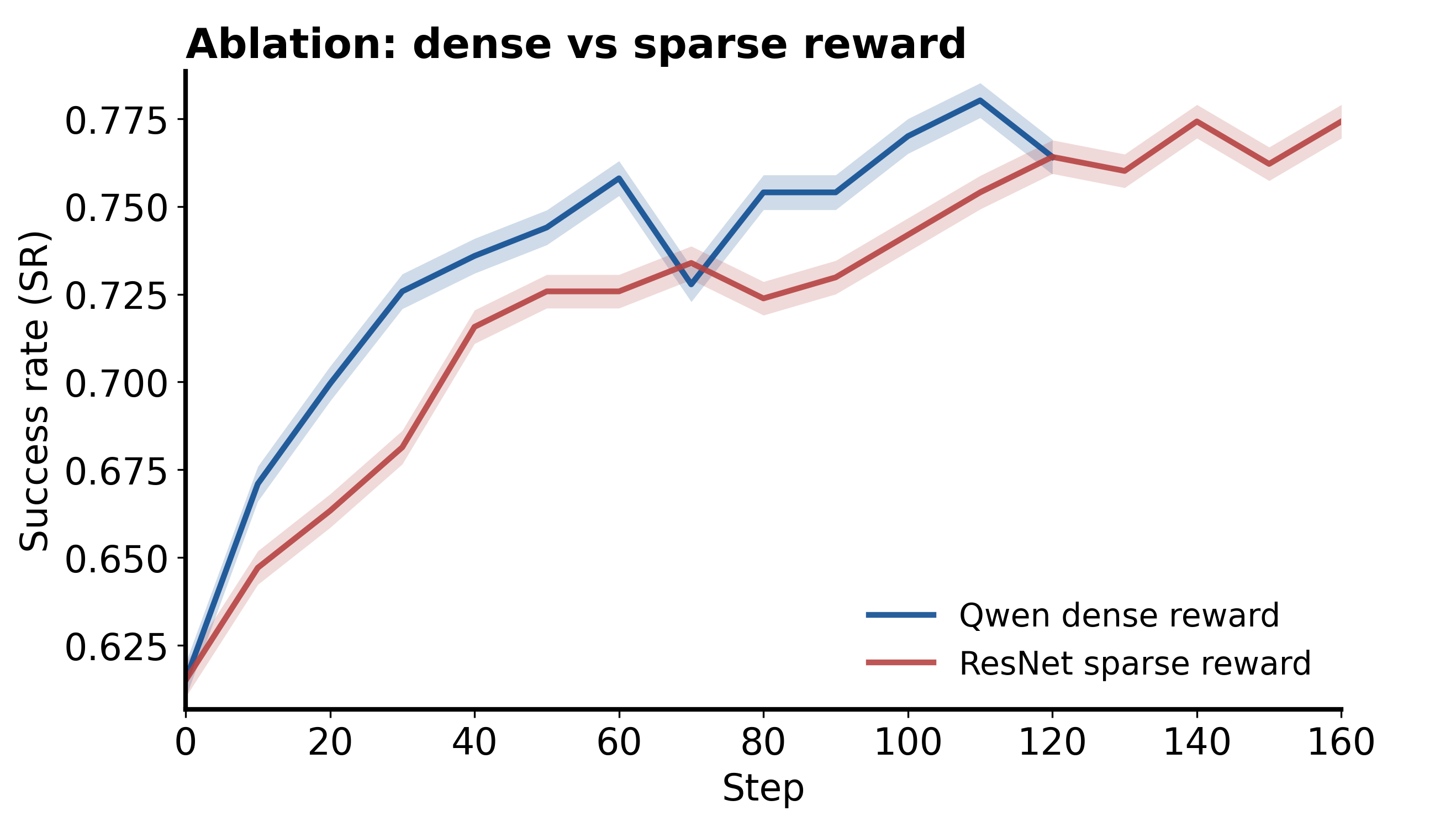}
    \caption{Ablation on reward modeling: dense vs. sparse reward.}
    \label{fig:ablation_reward}
\end{figure}

Fig.~\ref{fig:ablation_reward} shows the learning curves under the two reward settings. We observe that the dense reward model improves sample efficiency in the early stage of training, leading to faster initial performance gains. However, both reward designs converge to similar final performance.

Despite its advantage in early learning, the dense reward model incurs significantly higher computational cost, as it relies on a large vision-language backbone (Qwen3-VL 2B). In practice, we find that rollout with the dense reward model is approximately $3\times$ slower than with the lightweight sparse reward model. 

What's more, binary rewards are widely adopted in manipulation benchmarks such as LIBERO as well as real-world robotic setups. Taking the above accounts into considerations,  we adopt the sparse reward model in all main experiments.

\section{Qualitative Failure Mode Analysis}
\label{sec:fail_mode}

\begin{figure}[ht]
    \centering
    \includegraphics[width=0.99\linewidth]{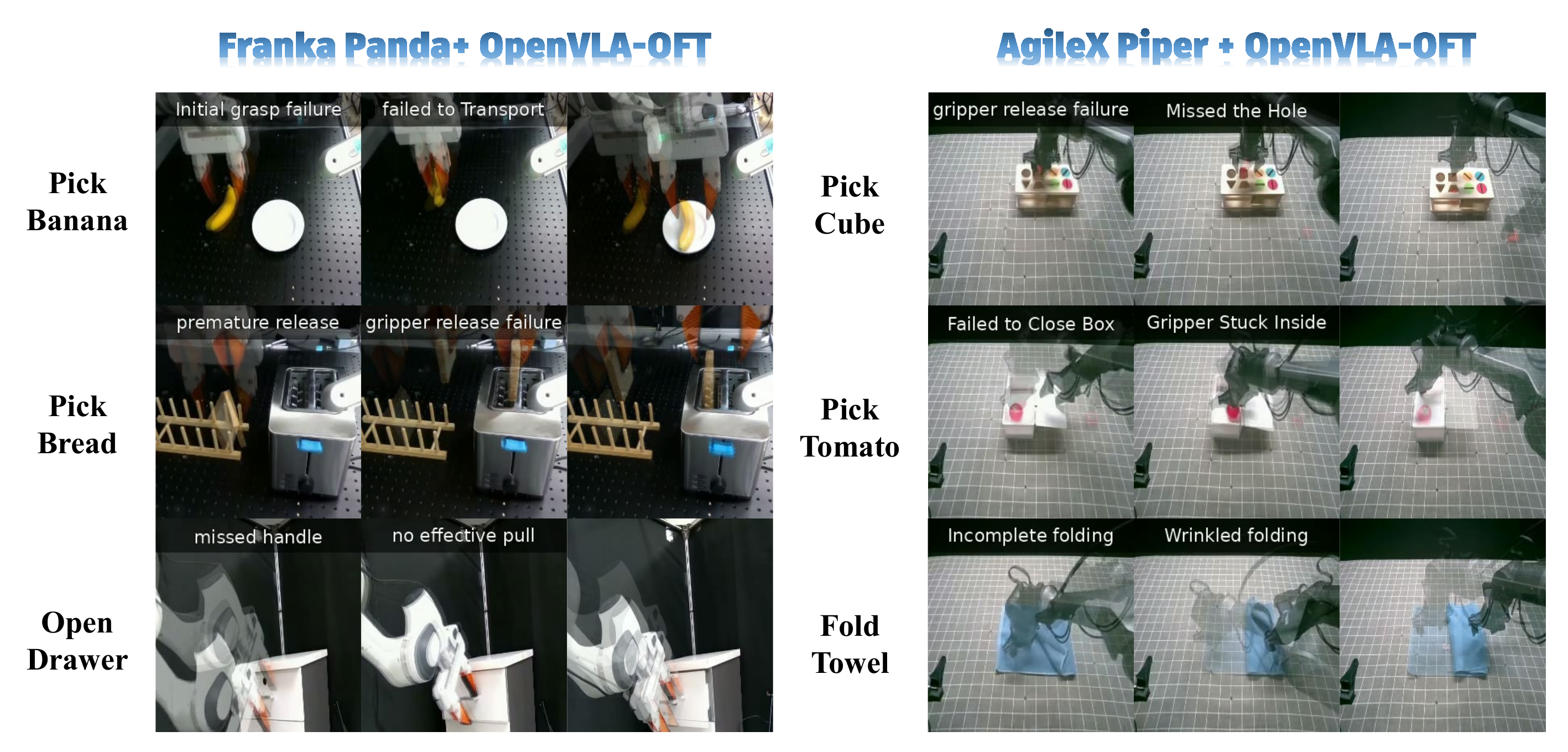}
    \caption{Visualization of real-world execution before and after WoVR policy optimization.}
    \label{fig:failure_analysis}
\end{figure}

We further analyze representative failure modes of base VLA model and compare them with the behaviors after WoVR. Fig.~\ref{fig:failure_analysis} visualizes execution traces across two robotic platforms and six manipulation tasks. The comparison shows that WoVR not only improves task success rates, but also changes the dominant failure patterns of the base VLA policy.

On the Franka Panda platform, the base policy exhibits different failure modes across the three tasks. In \textbf{Pick Banana}, the robot often repeatedly attempts to grasp the banana without establishing a stable grasp, or successfully grasps the banana but fails to move it above the plate. After applying WoVR, the policy more reliably grasps the banana and transfers it into the plate with faster and more direct motions. In \textbf{Pick Bread}, the base policy mainly fails by prematurely opening the gripper, causing the bread to drop, or by keeping the gripper closed after reaching the target region. WoVR largely suppresses these two failure modes and produces more consistent release behavior. In \textbf{Open Drawer}, the base policy frequently fails to reach the handle accurately or fails to pull the drawer after making contact. With WoVR, the robot more often completes the full interaction sequence, including handle approach, contact, and drawer pulling.

On the AgileX Piper platform, failures are more strongly affected by noisier state estimation and less stable low-level control. In \textbf{Pick Cube}, the base policy often moves the cube to the target region but fails to release the gripper, or places the cube with an incorrect pose such that it does not fall into the target hole. WoVR substantially reduces these errors by producing more stable placement and release behaviors. In \textbf{Pick Tomato}, the base policy commonly fails to close the box after placing the tomato, or leaves the gripper trapped inside the box. After WoVR optimization, these failure cases are mitigated, although the task remains sensitive to contact and box geometry. In \textbf{Fold Towel}, the base policy often fails to complete the fold or keeps holding the towel after folding. WoVR improves the temporal coordination between folding and release, leading to a higher success rate on this deformable-object task.

Overall, the qualitative analysis suggests that WoVR improves real-world execution by reducing recurrent action-level failure modes, including unstable grasping, premature or delayed gripper release, inaccurate target placement, and incomplete contact-rich interactions. These results are consistent with the quantitative real-world improvements reported in Sec.~\ref{sec:q3}, and indicate that hallucination-aware policy optimization in imagination can translate into more reliable closed-loop behavior on physical robots.

\section{Real-World Experiments}
\label{sec:real_world}

\subsection{Hardware Setup}
\label{app:rw:hardware}
\begin{figure}[t]
\centering
\begin{subfigure}[b]{0.48\textwidth}
\centering
\includegraphics[width=\linewidth]{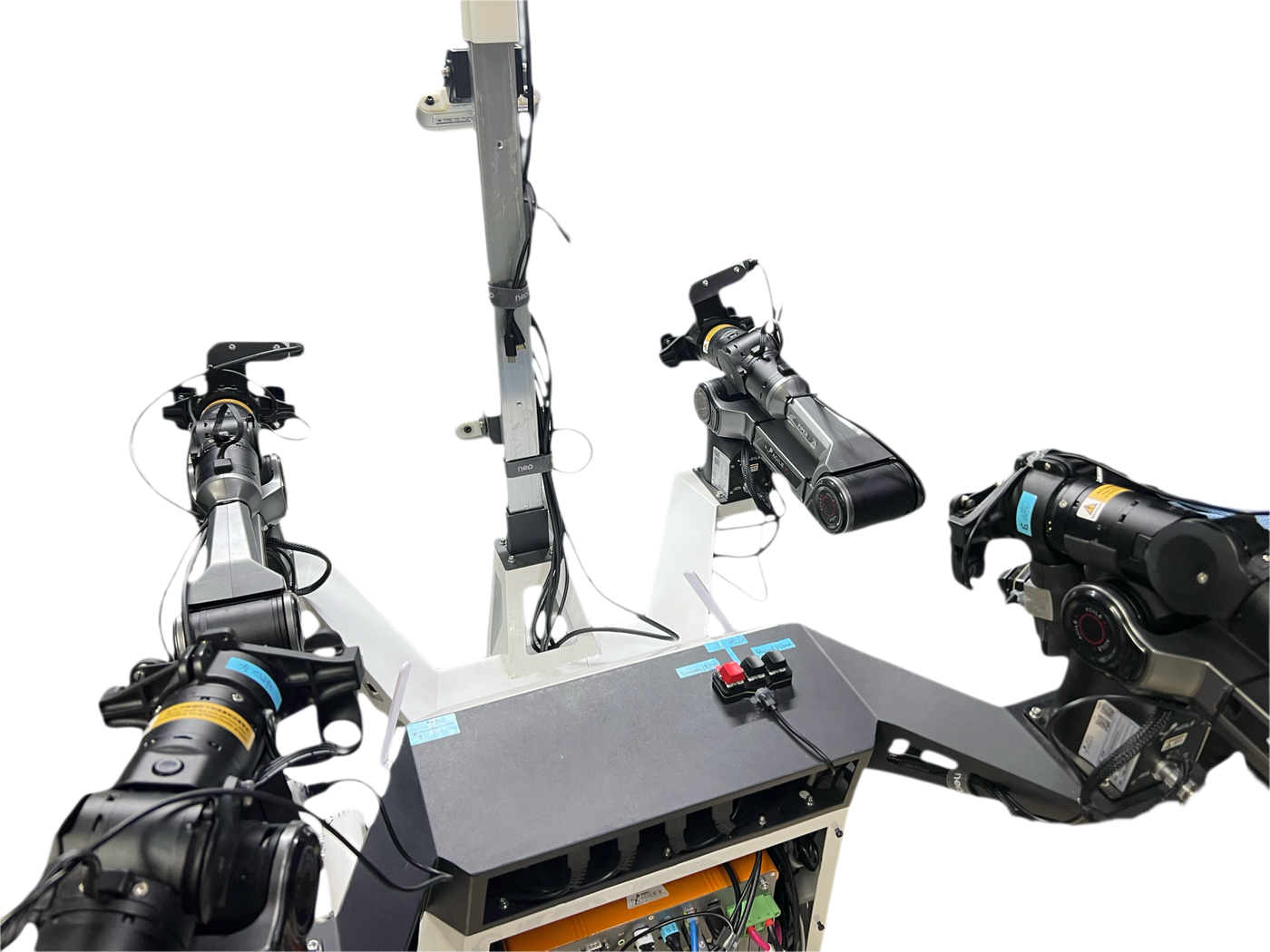}
\caption{AgileX Piper}
\end{subfigure}
\hfill  
\begin{subfigure}[b]{0.48\textwidth}
\centering
\includegraphics[width=\linewidth]{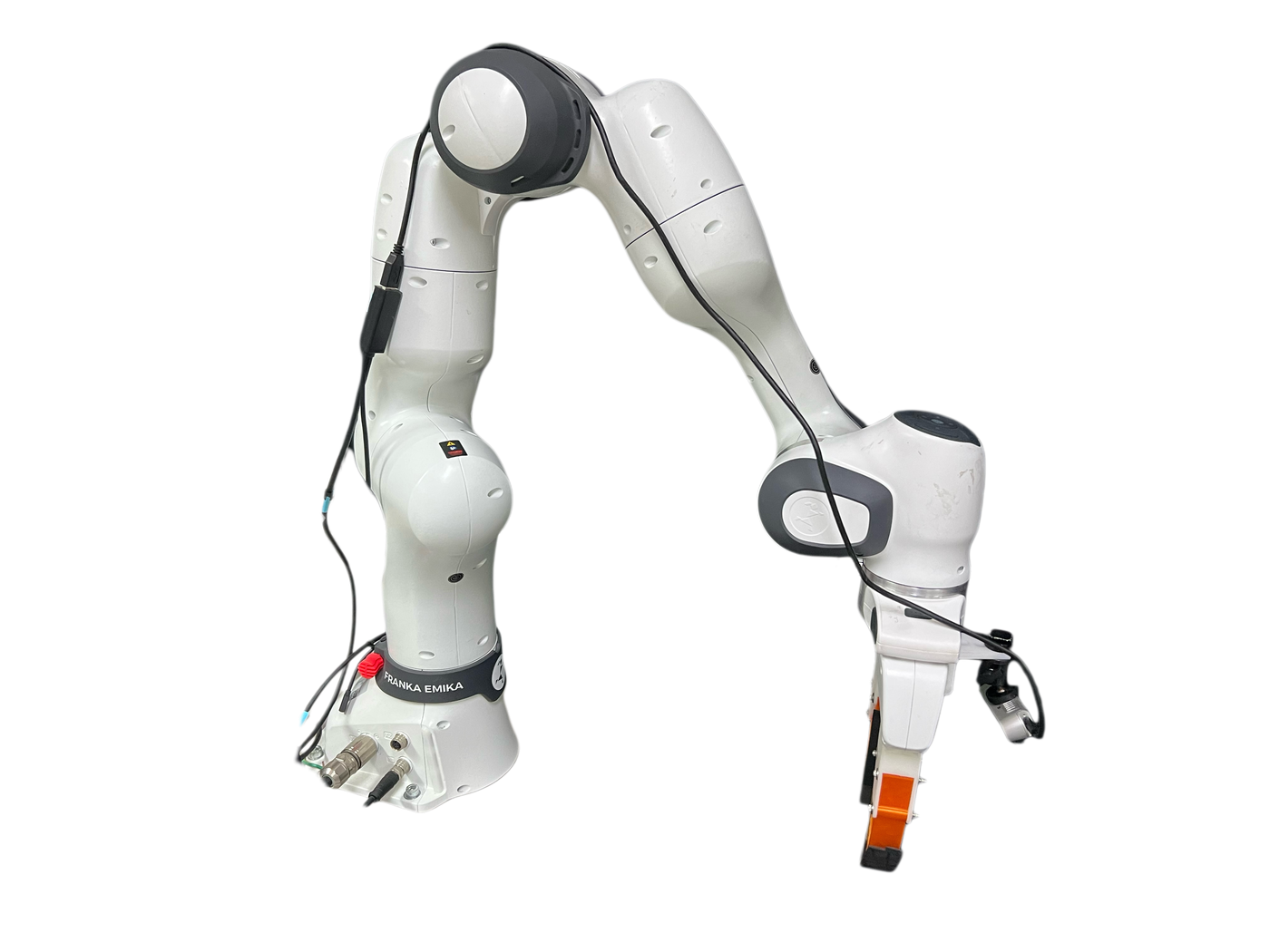}
\caption{Franka Emika Panda}
\end{subfigure}
\caption{Two types of robotic arm platforms}
\label{fig:robotic arm platforms}
\end{figure}

We evaluate our method on two distinct real-world robotic platforms: the AgileX Piper robotic arm (left) and the Franka Emika Panda robotic arm (right) as shown in Fig.\ref{fig:robotic arm platforms}, each equipped with two-finger grippers. This was done to test the robustness of our method under different control precision hardware conditions.

\paragraph{AgileX Piper}
AgileX Piper is a 6-degree-of-freedom robotic arm. We conduct remote operation based on the \textbf{cobot Magic} platform. We use the D435 camera with a head view to capture RGB images and synchronize and record the data at a frequency of 10Hz.

\paragraph{Franka Emika Panda}
Franka Emika Panda is a 7-degree-of-freedom robotic arm. We use the fixed third-person perspective D435 to record RGB images and use a spatial mouse for remote operation. Data is recorded at a frequency of 10Hz.

\subsection{Detailed task requirements}
\label{app:rw:detailed}

Here, we provide the detailed requirements for each task on both robotic platforms.

On Franka platform, we evaluate three manipulationtasks:
\begin{itemize}
    \item Pick Banana: Pick a banana and place it onto a plate;
    \item Pick Bread: Pick a bread and place it into the toaster;
    \item Open Drawer: Pull the drawer open;
\end{itemize}

On AgileX Piper platform,we also evaluate three other tasks: 
\begin{itemize}
    \item Pick Cube: Pick up a cube and place it into a box;
    \item Pick Tomato: Pick up a tomato, place it into a box, and close the lid;
    \item Fold Towel: Fold the towel into a compact configuration.
\end{itemize}

\subsection{Practical Considerations for Real-World Deployment}
\label{app:rw:practical}

Here, we introduce the selection of the action space and control method for the robotic arm. This is of crucial importance for the actual operation of the less precise robotic arm.

\paragraph{Platforms and Action Space.}
Experiments are conducted on two robotic platforms: the \textbf{AgileX Piper} manipulator and the \textbf{Franka Emika Panda}. For both platforms, the action space is defined in the delta end-effector (delta EEF) space, where the VLA policy outputs relative end-effector displacements that are also used as inputs to the world model.

Since the lower-level drives of both robotic arms are controlled in the absolute joint space, each predicted delta action is converted into an absolute end-effector pose, followed by inverse kinematics (IK) to obtain executable joint commands.

\paragraph{Rotation Parameterization.}
A critical design choice is the parameterization of rotational actions. Directly regressing $\Delta$RPY is often unstable, since Euler angles suffer from discontinuities and singularities, and small prediction errors may lead to large deviations in the reconstructed absolute orientation, thereby affecting IK consistency.

To improve robustness, we adopt the Rotation 6D representation~\citep{zheng2025xvlasoftpromptedtransformerscalable,zhou2020continuityrotationrepresentationsneural}. 
Given the current end-effector orientation $R_t \in SO(3)$ and the next orientation $R_{t+1} \in SO(3)$, we first compute the local relative rotation:
\begin{equation}
    \Delta R_t = R_t^{\top} R_{t+1}.
\end{equation}
Following the standard 6D rotation representation, $\mathrm{RotMatTo6D}(\cdot)$ denotes flattening the first two columns of a rotation matrix. The corresponding delta rotation in 6D form is computed as:
\begin{equation}
    \Delta \mathbf{r}_{6D,t}
    =
    \mathrm{RotMatTo6D}(\Delta R_t)
    =
    \left[
    \Delta R_t[:,1]^\top,
    \Delta R_t[:,2]^\top
    \right]^\top
    \in \mathbb{R}^{6}.
\end{equation}
Here, $\Delta \mathbf{r}_{6D,t}$ represents the relative rotation from the current pose to the next pose in the local end-effector frame.

During execution, the predicted 6D vector is converted back to a valid rotation matrix through Gram--Schmidt orthogonalization. Specifically, given two predicted vectors $\mathbf{a}_1,\mathbf{a}_2 \in \mathbb{R}^{3}$, we compute:
\begin{equation}
    \mathbf{b}_1 = \frac{\mathbf{a}_1}{\|\mathbf{a}_1\|},
\end{equation}
\begin{equation}
    \mathbf{b}_2 =
    \frac{
    \mathbf{a}_2 - (\mathbf{b}_1^\top \mathbf{a}_2)\mathbf{b}_1
    }{
    \left\|
    \mathbf{a}_2 - (\mathbf{b}_1^\top \mathbf{a}_2)\mathbf{b}_1
    \right\|
    },
\end{equation}
\begin{equation}
    \mathbf{b}_3 = \mathbf{b}_1 \times \mathbf{b}_2,
    \qquad
    \widehat{\Delta R}_t =
    [\mathbf{b}_1,\mathbf{b}_2,\mathbf{b}_3].
\end{equation}
The absolute orientation is then updated as:
\begin{equation}
    \widehat{R}_{t+1} = R_t \widehat{\Delta R}_t .
\end{equation}

Compared with RPY, Rotation 6D avoids angle wrapping and gimbal-lock singularities. Compared with quaternions, it avoids the double-cover ambiguity, where $\mathbf{q}$ and $-\mathbf{q}$ represent the same rotation. In addition, the network can directly regress an unconstrained 6D vector, which is later projected to a valid rotation matrix by orthogonalization. This makes the action representation more continuous and better suited for learning stable end-effector rotation commands.

The final action space is defined as:
\[
(\Delta x, \Delta y, \Delta z, \Delta \mathbf{r}_{6D}, \text{gripper}).
\]

\paragraph{Platform-Specific Considerations.}
While the overall control pipeline is shared, the two platforms exhibit different levels of state estimation accuracy.

The Franka Panda provides highly accurate and stable end-effector state feedback, making the standard update
\[
\mathbf{a}_i = \mathbf{s}_i + \Delta \mathbf{a}_i
\]
sufficient in practice. Here, the addition only denotes delta-to-absolute action conversion, not direct physical motion composition.

In contrast, the AgileX Piper exhibits noticeable noise and bias in the observed end-effector state, which leads to drift when applying the same update rule. To mitigate this issue, we instead compute actions using an accumulated formulation:
\[
\mathbf{a}_i = \mathbf{s}_0 + \sum_{k=1}^{i} \Delta \mathbf{a}_k.
\]
Here, $s_{0}$ represents the given initial position. This reduces sensitivity to noisy feedback and improves long-horizon execution stability.

Although this formulation may introduce discrepancies under external disturbances (e.g., contacts), such errors are compensated by the visual feedback loop of the VLA policy.

\paragraph{Summary.}
These results highlight that while the proposed delta-space formulation generalize across platforms, handling state estimation noise is critical for reliable real-world deployment. The proposed design enables stable closed-loop execution under both high-precision and low-cost hardware settings.

\subsection{Task Setting}
\label{app:rw:task}

\begin{table*}[t]
\centering
\caption{Task and embodiment settings in WoVR real-world experiments. 
Task-specific data scales are reported in Panel (a), while embodiment-specific training configurations are summarized in Panel (b).}
\label{tab:wovr_real_robot_settings}
\renewcommand{\arraystretch}{1.15}
\setlength{\tabcolsep}{6pt}
\normalsize

\textbf{(a) Task-specific data scale.}\\[3pt]
\begin{tabular}{llcc}
\toprule
Embodiment 
& Task 
& VLA SFT Demos 
& WM SFT Rollouts \\
\midrule
Franka 
& Pick banana 
& 25 
& 120 \\

Franka 
& Pick bread 
& 25 
& 120 \\

Franka 
& Open drawer 
& 25 
& 120 \\
\midrule
Agilex 
& Pick cube 
& 75 
& 180 \\

Agilex 
& Pick tomato 
& 50 
& 180 \\

Agilex 
& Fold towel 
& 50 
& 210 \\
\bottomrule
\end{tabular}

\vspace{0.8em}

\textbf{(b) Embodiment-specific training configuration.}\\[3pt]
\begin{tabular}{lcccc}
\toprule
Embodiment 
& VLA SFT Steps 
& WM SFT Steps 
& Max Episode Steps 
& RL Epochs \\
\midrule
Franka 
& 50k 
& 750k 
& 160 
& 50 \\

Agilex 
& \makecell{50k (OpenVLA-OFT)\\30k ($\pi_{0.5}$)}
& 1.25M 
& 224 
& 60 \\
\bottomrule
\end{tabular}

\end{table*}

The amount of VLA SFT data differs across tasks because the tasks have different levels of difficulty. We allocate more demonstrations to harder tasks to ensure that the supervised policy reaches a sufficient initial performance before reinforcement learning. The amount of data used for world model training follows the same principle: tasks with more complex dynamics, longer horizons, or more diverse interactions require more rollout data for reliable modeling.

In our real-world experiments, we do not use PACE for data collection due to efficiency considerations. Instead, we construct the world-model training set by rolling out multiple VLA SFT checkpoints saved at different training steps. This strategy increases trajectory diversity and broadens action-space coverage, as checkpoints from different training stages induce distinct behavior distributions. For example, on the Franka platform, we collect rollout trajectories from checkpoints saved at 25k, 30k, 35k, 40k, 45k, and 50k SFT steps. We then select the best-performing checkpoint based on real-world evaluation, report its performance, and use it as the initialization for subsequent RL.

\end{document}